\documentclass{article}

\PassOptionsToPackage{numbers, compress}{natbib}
\usepackage[preprint]{neurips_2023}

\usepackage{amsmath,amsfonts,bm}

\def\eqref#1{equation~\ref{#1}}

\def\1{\bm{1}}

\def\va{{\bm{a}}}
\def\vb{{\bm{b}}}
\def\vc{{\bm{c}}}
\def\vd{{\bm{d}}}
\def\ve{{\bm{e}}}

\def\vh{{\bm{h}}}

\def\vr{{\bm{r}}}

\def\vx{{\bm{x}}}
\def\vy{{\bm{y}}}
\def\vz{{\bm{z}}}

\def\mH{{\bm{H}}}

\def\mR{{\bm{R}}}

\def\mX{{\bm{X}}}
\def\mY{{\bm{Y}}}
\def\mZ{{\bm{Z}}}

\DeclareMathAlphabet{\mathsfit}{\encodingdefault}{\sfdefault}{m}{sl}
\SetMathAlphabet{\mathsfit}{bold}{\encodingdefault}{\sfdefault}{bx}{n}

\def\gE{{\mathcal{E}}}

\def\gG{{\mathcal{G}}}

\def\gL{{\mathcal{L}}}

\def\gO{{\mathcal{O}}}

\def\gR{{\mathcal{R}}}

\def\gV{{\mathcal{V}}}

\def\gZ{{\mathcal{Z}}}

\def\tta{{\texttt{a}}}
\def\ttb{{\texttt{b}}}
\def\ttc{{\texttt{c}}}
\def\ttd{{\texttt{d}}}
\def\tte{{\texttt{e}}}

\def\ttr{{\texttt{r}}}

\def\ttz{{\texttt{z}}}

\usepackage[utf8]{inputenc} 
\usepackage[T1]{fontenc}    
\usepackage{hyperref}
\usepackage{url}
\usepackage{booktabs}       
\usepackage{amsfonts}       
\usepackage{nicefrac}       
\usepackage{microtype}      
\usepackage{xcolor}         
\usepackage{amsmath}
\usepackage{amssymb}
\usepackage{mathtools}
\usepackage{amsthm}
\usepackage{makecell}
\usepackage{stackrel}
\usepackage{enumitem}
\usepackage{multirow}
\usepackage{hyperref}
\usepackage{cleveref}
\usepackage{wrapfig}
\hypersetup{
    colorlinks = true,
    citecolor = blue,
    linkcolor = red
}

\usepackage{graphicx,subfigure}
\usepackage{caption}
\definecolor{red}{RGB}{205,33,42}
\definecolor{blue}{RGB}{33,105,225}
\definecolor{pptblue}{RGB}{67,114,196}
\definecolor{amethyst}{rgb}{0.6, 0.4, 0.8}
\definecolor{applegreen}{rgb}{0.55, 0.71, 0.0}
\definecolor{realblue}{RGB}{0,0,255}

\DeclarePairedDelimiterX\set[2]\{\}{#1\mathrel{}\mathclose{}\delimsize|\mathopen{}\mathrel{}#2}

\theoremstyle{plain}
\newtheorem{theorem}{Theorem}[section]
\newtheorem{proposition}[theorem]{Proposition}

\theoremstyle{definition}
\newtheorem{definition}[theorem]{Definition}

\theoremstyle{remark}

\theoremstyle{definition}
\newtheorem{innercustomgeneric}{\customgenericname}
\providecommand{\customgenericname}{}
\newcommand{\newcustomtheorem}[2]{
  \newenvironment{#1}[1]
  {
   \renewcommand\customgenericname{#2}
   \renewcommand\theinnercustomgeneric{##1}
   \innercustomgeneric
  }
  {\endinnercustomgeneric}
}
\newcustomtheorem{customproposition}{Proposition}

\title{Knowledge Graph-Augmented Language Models\\ for Knowledge-Grounded Dialogue Generation}

\author{
  Minki Kang$^{1,2}$\thanks{Equal Contribution.},\quad
  Jin Myung Kwak$^{1*}$,\quad
  Jinheon Baek$^{1*}$,\quad
  \textbf{Sung Ju Hwang}$^{1}$ \\
  KAIST$^{1}$, AITRICS$^{2}$ \\
  \texttt{\{zzxc1133, kwak.jinmyung, jinheon.baek, sjhwang82\}@kaist.ac.kr}
}

\begin{document}

\maketitle

\begin{abstract}
Language models have achieved impressive performances on dialogue generation tasks. However, when generating responses for a conversation that requires factual knowledge, they are far from perfect, due to an absence of mechanisms to retrieve, encode, and reflect the knowledge in the generated responses. Some knowledge-grounded dialogue generation methods tackle this problem by leveraging facts from Knowledge Graphs (KGs); however, they do not guarantee that the model utilizes a relevant piece of knowledge from the KG. To overcome this limitation, we propose \textbf{SU}bgraph \textbf{R}etrieval-augmented \textbf{GE}neration (\textbf{SURGE}), a framework for generating context-relevant and knowledge-grounded dialogues with the KG. Specifically, our SURGE framework first retrieves the relevant subgraph from the KG, and then enforces consistency across facts by perturbing their word embeddings conditioned by the retrieved subgraph. Then, we utilize contrastive learning to ensure that the generated texts have high similarity to the retrieved subgraphs. We validate our SURGE framework on OpendialKG and KOMODIS datasets, showing that it generates high-quality dialogues that faithfully reflect the knowledge from KG. 
\end{abstract}
\section{Introduction}
\label{sec:intro}
\begin{figure*}
    \centering
    \includegraphics[width=0.95\linewidth]{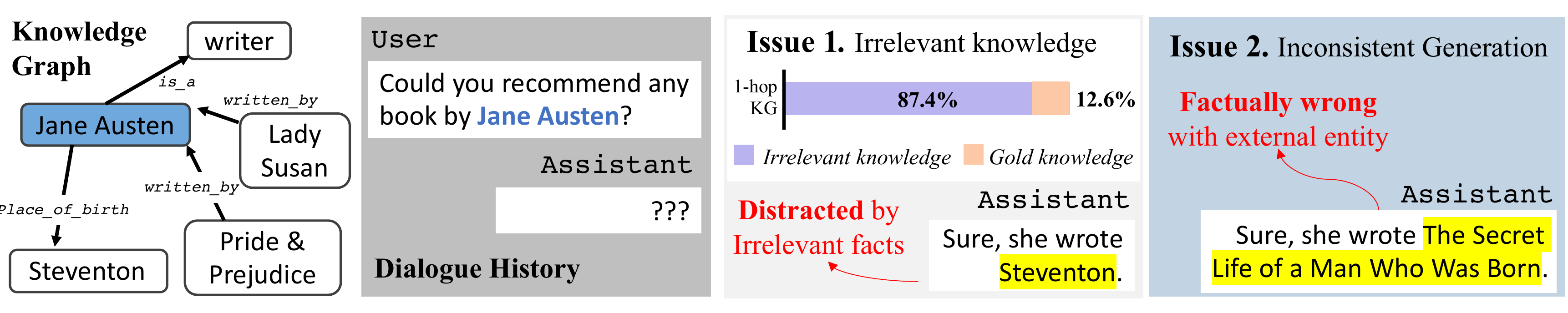}
    \vspace{-0.05in}
    \caption{\small \textbf{Motivation.} Existing knowledge-grounded dialogue generation models with a KG often utilize the multi-hop subgraph associated with entities in the dialogue context (e.g., \textbf{\textcolor{pptblue}{Jane Austen}}). However, they suffer from a couple of the following problems: \textbf{(1) irrelevant knowledge} where only 12.6\% of facts from the 1-hop KG are useful to generate the target responses given a dialogue context, and \textbf{(2) inconsistent generation} where generated texts include the factually incorrect statements.}
    \vspace{-0.1in}
    \label{fig:concept}
\end{figure*}

\vspace{-0.05in}
Dialogue systems aim to engage in ongoing conversations with humans by generating human-like responses. While pre-trained language models (PLMs)~\citep{GPT2, T5} are capable of generating fluent responses, they often generate factually incorrect responses due to a lack of explicit knowledge~\citep{RAG-FiD}. 
To overcome such limitations, recent methods access external knowledge sources such as Wikipedia~\citep{WoW}, Web~\citep{WoI}, Freebase  ~\citep{Freebase} and Wikidata~\citep{wikidata} to retrieve the relevant knowledge for the dialogue context. 
In this work, we focus on the Knowledge Graphs (KGs)-based dialogue generation as existing works~\citep{DyKgChat,75s5a,75s5b,75s5c,EARL,efficientdialogue,PLUG}. KGs represent facts in the most compact and effective symbolic structured form (See the leftmost of \autoref{fig:concept}), consisting of entities as nodes and their relation as an edge. Each of them is formed with a triplet, which can help generate knowledge-grounded responses.

Most of the dialogue generation models with KGs~\citep{efficientdialogue, PLUG} utilize all the triplets associated with entities in the dialogue context. However, through observation, we found not all of the facts are actually relevant to the conversation (e.g., Jane Austen was born in Steventon in \autoref{fig:concept}), which could mislead the models to generate factually incorrect responses.
87\% of facts from 1-hop KG are irrelevant to the context in the OpendialKG dataset~\citep{opendialkg}. Moreover, encoding all the facts including the unnecessary ones is computationally inefficient~\citep{efficientdialogue, DialoKG}.
On the other hand, even after correctly retrieving the relevant facts, it is not straightforward to combine two heterogeneous modalities: the dialogue context is represented as a \textit{text}; the knowledge is represented as a \textit{graph}.
In other words, since PLMs already have tons of pre-trained parameters trained on the unstructured texts, properly conditioning the structured graph to PLMs is highly challenging. Otherwise, PLMs may generate inconsistent responses disregarding the knowledge from the retrieved subgraph, which is a  phenomenon known as hallucination~\citep{Hallucination2}, where they generate responses with their own memorized yet incorrect knowledge.

In this work, we tackle such challenging and fundamental issues of knowledge-grounded dialogue generation with the KGs. We propose an end-to-end dialogue generation framework that considers all aspects from knowledge retrieval, encoding, and reflection along the generation process.   
As a first step, we propose a context-relevant subgraph retriever that retrieves only the relevant triplets from  KGs to prevent the model from generating context-irrelevant responses. 
Notably, our subgraph retrieval method embeds the KG considering it relational structure with the Graph Neural Network (GNN)~\citep{GCN} instead of using PLMs as in previous work~\citep{PLUG}.
Furthermore, without any labels for pairs of dialogue contexts and their relevant subgraphs, our method is end-to-end trainable jointly with the generation objective by marginalizing the likelihood of the generated sentences over the latent retrieved subgraph~\citep{REALM,RAG}.
Then, to encode the retrieved subgraph along with the input text sequence, we propose a graph encoding that is permutation and relation inversion invariant yet efficient. Specifically, we devise the graph encoding method that reflects the graph structure onto the representation space of PLMs, instead of prepending them in front of the text sequence to avoid the computational burden.
Furthermore, to ensure that the model does make use of the encoded knowledge when generating responses, we propose a multi-modal contrastive learning objective between two different graph-text modalities to enforce consistency across the retrieved facts and the generated texts. 
We call our framework \textbf{SU}bgraph \textbf{R}etrieval-augmented \textbf{GE}neration (\textbf{SURGE}).

We validate our SURGE framework on the OpendialKG~\citep{opendialkg} and KOMODIS~\citep{KOMODIS} datasets.
Note that, when evaluating the generated responses from dialogue models, conventional metrics (e.g., BLEU~\citep{BLEU}; Rouge~\citep{ROUGE}) can not measure how faithfully the generated responses reflect the related knowledge in KGs. 
Thus, in evaluation, we further introduce an additional performance metric, referred to as Knowledge-verifying Question Answering (KQA), which evaluates whether the generated responses contain the correct knowledge with an additional extractive question answering scheme. The experimental results show that SURGE generates responses that not only agree with the gold knowledge but are also consistent with the retrieved knowledge from KGs. Our contributions are summarized as follows:
\vspace{-0.075in}
\begin{itemize} [itemsep=0.5mm, parsep=1pt, leftmargin=*]
    \item We propose a GNN-based context-relevant subgraph retriever that extracts the context-relevant piece of knowledge from KGs, for generating appropriate responses to the ongoing conversation.
    \item We propose an invariant yet efficient graph encoder and a graph-text contrastive learning objective to ensure that the generated responses faithfully reflect the retrieved knowledge.  
    \item We validate our SURGE framework against relevant baselines, demonstrating its efficacy in generating responses that are more informative by retrieving and reflecting the relevant knowledge.
\end{itemize}
\section{Related Work}

\paragraph{Language Models}
Pre-trained Language Models (PLMs)~\citep{GPT2, BART, T5} that use a Transformers-based~\citep{transformer} encoder-decoder architecture have achieved great successes on language generation tasks. As they can accurately contextualize the given context and then generate human-like sentences, they are recently used as the base architecture for neural dialogue systems~\citep{zhang2019dialogpt, SimpleTOD}. Moreover, when PLMs become larger, dialogue models are capable of generating high-quality responses~\citep{Meena}, suggesting that pre-trained parameters do contain certain knowledge~\citep{KGLM}. However, despite the fluency of such dialogue methods, they often generate factually incorrect responses that are unfaithful to the context but look plausible -- widely known as hallucination~\citep{Hallucination}. 
To tackle this challenge, recent studies~\citep{Blenderbot, RAG-FiD} propose to retrieve knowledge from external sources, and then augment it to dialogue models.

\paragraph{Dialogue Generation with KGs}
Regarding dialogue generation tasks with KGs that we target, \citet{opendialkg} introduce a knowledge-grounded dialogue dataset, where each dialogue turn comes with facts from the large-scale KG.
Several following works~\citep{DyKgChat,75s5a,75s5b,75s5c,EARL} suggest sequence-to-sequence models trained from scratch, which focus on generating dialogue by conditioning the output word distribution with entities from the KG.
Further, \citet{efficientdialogue} propose an efficient method that encodes all facts in the $k$-hop neighbors of entities that appear in the dialogue history, in order to reduce the number of input tokens forwarded in PLMs.
On the other hand, \citet{DialoKG} propose to mask out model weights for irrelevant facts in PLMs.
However, all of these methods simply match and retrieve \textit{all facts} for entities that appear in the dialogue context, which either may mislead models to generate out-of-context responses from irrelevant facts, or can increase the computational overheads from prepending all tokens for all facts in PLMs.
Our work differs from them since we aim at retrieving only a context-relevant subgraph among all associated facts with its retriever, which is end-to-end trainable along with a generative model.

\section{Method}
\label{sec:method}

\begin{figure*}
    \centering
    \includegraphics[width=0.99\linewidth]{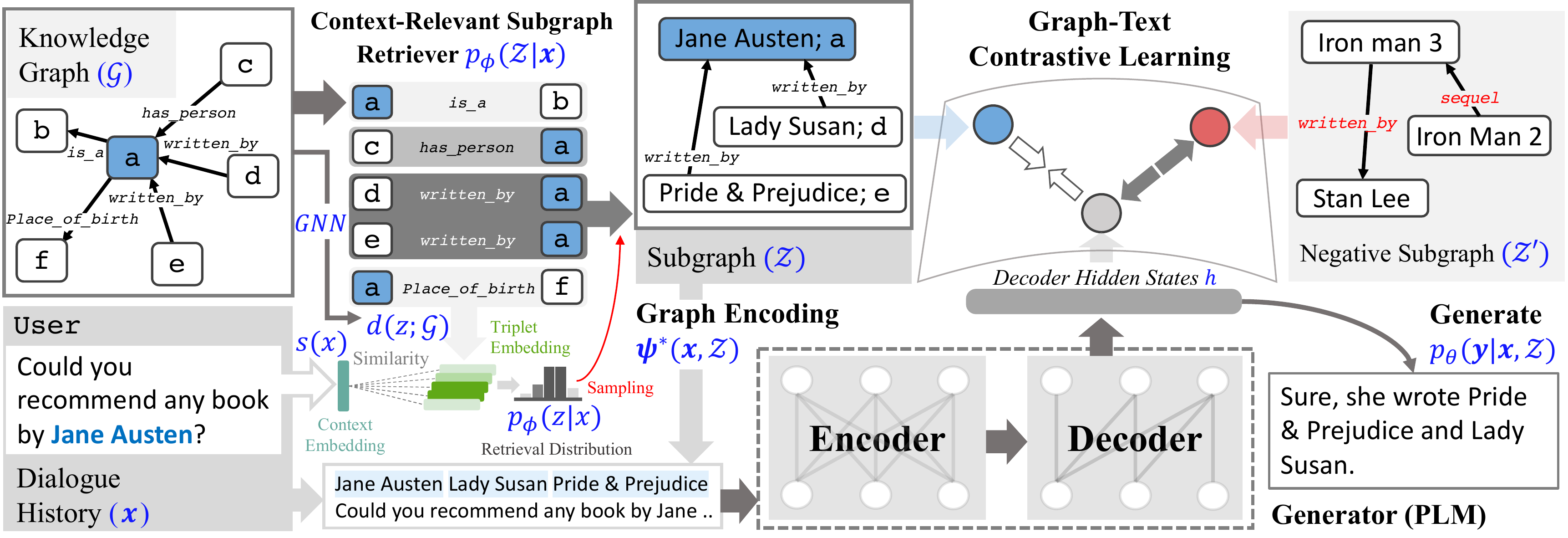}
    \caption{\small \textbf{Framework Overview.} Our framework, SURGE, consists of three parts. First, a context-relevant subgraph retriever $p_\phi (\gZ | \vx)$ retrieves the subgraph $\gZ$ relevant to the given dialogue history $\vx$ from a knowledge graph $\gG$ (e.g., 1-hop KG from entity \textit{Jane Austen}; \tta). 
    Specifically, we measure the similarity of a context and triplet embedding to compose the retrieval distribution $p_\phi (z|\vx)$ (\S~\ref{sec:retriever}).
    Then, we encode the retrieved subgraph $\gZ$ using the graph encoding $\boldsymbol{\psi}(\vx, \gZ)$ (\S~\ref{sec:encoding}).
    Finally, we use contrastive learning to enforce the model to generate a knowledge-grounded response with the retrieved subgraph (\S~\ref{sec:contrastive}).}
    \label{fig:framework}
\end{figure*}

In this section, we first discuss the basic ingredients: Transformer and Graph Neural Network. We then formalize the dialogue generation problem and describe key components for our \textbf{SU}bgraph \textbf{R}etrieval-augmented \textbf{GE}neration (\textbf{SURGE}) framework: context-relevant subgraph retrieval, invariant graph encoding, and graph-text contrastive learning. \autoref{fig:framework} illustrates the overview of our framework.

\subsection{Preliminaries}
\label{sec:prelim}
As we use two different modalities, namely text and graph, we first define them, and then describe the neural networks to encode them. In particular, a text is defined as a sequence of tokens $\vx = [x_1, ..., x_N], \forall x_i \in \mathcal{V}$, where $x_i$ is a token and $\gV$ is a pre-defined vocabulary formed with specific tokenization algorithms~\citep{BPE}. On the other hand, a knowledge graph (KG) is a type of multi-relational graphs $\mathcal{G} = \{(\tte_h, \ttr, \tte_t)\} \in \gE \times \gR \times \gE $, where $\tte_h$ and $\tte_t$ are head and tail entities (nodes) along with their relation (edge) $\ttr$; and $\mathcal{E}$ and $\mathcal{R}$ are sets of entities and relations, respectively.

To easily access different modalities in the same framework, we define the tokenization (mapping) function that maps entities and relations to tokens used in Pre-trained Language Models (PLMs), represented as follows: $q:\mathcal{E} \cup \mathcal{R} \rightarrow \mathcal{V}^l$ where $l$ is an arbitrary length varying across different entities and relations. 
In other words, any entity $\tte \in \mathcal{E}$ and relation $\ttr \in \mathcal{R}$ is tokenized to a sequence of $l$ tokens $\vx \in \gV^l$: $q(\tte) = \vx_e$ and $q(\ttr) = \vx_r$. 
For instance, an entity \textit{New York} (i.e., $\tte$), is tokenized into two tokens `New' and `York', i.e., $\vx_e = [\text{`New'}, \text{`York'}]$.

\paragraph{Transformer}
A Transformer~\citep{transformer} is the most basic building block of recent PLMs~\citep{BERT, GPT2}. 
Given a sequence $\vx = [x_1, ..., x_N], \forall x_i \in \mathcal{V}$, generative transformers generate a sequence $\vy_{1:t-1} = [y_1, ..., y_{t-1}], \forall y_i \in \mathcal{V}$, with encoder \texttt{Enc}, decoder \texttt{Dec}, and token embedding $f$. A hidden state at time $t$ for generating $y_t$ is $\vh_t = \texttt{Dec}(\texttt{Enc}(\mX), \mY_{1:t-1})$, where $\mX = f(\vx) = [f(x_1), ..., f(x_N)]$ and $\mY_{1:t-1} = f(\vy_{1:t-1}) = [f(y_1), ..., f(y_{t-1})]$.
Both \texttt{Enc} and \texttt{Dec} functions are \textbf{permutation sensitive} with positional embedding.

\paragraph{Graph Neural Network}
\label{sec:gnn}
A Graph Neural Network (GNN) represents a node with its neighboring nodes over graphs~\citep{GRLbook}, as follows:
\begin{equation}
\fontsize{9pt}{9pt}\selectfont
    \texttt{GNN}(\ve_t;\gG) \nonumber = \texttt{UPD} (\ve_t, \texttt{AGG} (\set{\ve_h}{\forall \tte_h \in \mathcal{N}(\tte_t;\mathcal{G})} )),
\fontsize{9pt}{9pt}\selectfont
\label{eqn:gnn}
\end{equation}
where $\mathcal{N}(\tte_t;\mathcal{G}) = \set{\tte_h}{(\tte_h,\ttr,\tte_t)\in\mathcal{G}}$ is a set of neighboring entities of $\tte_t$; $\ve_t$ and $\ve_h$ are embeddings of entities (nodes) $\tte_t$ and $\tte_h$; \texttt{AGG} is a function that aggregates embeddings of neighboring entities; and \texttt{UPD} is a function that updates $\ve_t$ with the aggregated messages from \texttt{AGG}.

\subsection{Problem Statement}
Given a dialogue history $\vx = [x_1, \ldots, x_N]$, a model with generative PLMs models a conditional distribution $p(\vy|\vx)$ to generate an output response $\vy = [y_1, \ldots, y_T]$. 
To generate knowledge-grounded dialogue, this problem requires a piece of specific knowledge for an ongoing conversation. 

To that end, given a dialogue history $\vx$, we aim at retrieving a subgraph $\gZ \subseteq \gG$ consisting of a set of triplets $z \in \gZ$ where $z = (\tte_h, \ttr, \tte_t)$, which encodes relevant knowledge for ongoing conversation. Thus, the distribution of the context-relevant facts $\gZ$ is $p(\gZ|\vx)$, and our final likelihood of generating responses then becomes $p(\vy|\vx,\gZ)$. 
To jointly optimize the objective of graph retrieval with response generation, we treat $\gZ$ as a latent variable and then marginalize the likelihood of the generative model over all possible latent variables for retrieved subgraphs $\gZ$, formalized as follows:
\begin{equation}
\label{eqn:likelihood}
    p(\vy|\vx) = \sum_{\gZ \subseteq \gG} p_\phi (\gZ|\vx) \: p_\theta (\vy|\vx,\gZ) 
           = \sum_{\gZ \subseteq \gG} p_\phi (\gZ|\vx) \prod_{t=1}^T p_\theta (y_t|\vx,\gZ,\vy_{0:t-1}),
\end{equation}
where $y_0$ is the start token for the generation, $p_\phi(\gZ|\vx)$ is an output distribution of the context-relevant subgraph retriever, and $p_\theta(\vy|\vx,\gZ)$ is the target distribution of the knowledge-augmented generator, parameterized as $\phi$ and $\theta$, respectively, which we specify in next few subsections.

\subsection{Context-Relevant Subgraph Retriever}
\label{sec:retriever}
We now provide a concrete description of our context-relevant subgraph retriever, i.e., $p(\gZ|\vx)$, formalized in~\autoref{eqn:likelihood}. 
Given the dialogue history $\vx$, we assume that retrieval of each triplet in $\gZ = \{z_1, \ldots, z_n\}$ is independent. Then, for simplicity, we decompose the retrieval of a set of triplets $p(\gZ|\vx)$ into the product of individual triplet retrieval, represented as follows: $p(\gZ|\vx) = p(z_1|\vx)p(z_2|\vx) \cdots p(z_n|\vx)$, for $n$ retrieved triples. From this decomposition, it is now sufficient to focus on a single triplet retrieval. We define the score for the single triplet with the inner product between embeddings of the dialogue history $\vx$ and the candidate triplet $z$~\citep{REALM}, as follows:
\begin{equation}
    p_\phi(z|\vx) \propto \exp (d(z)^\top s(\vx)),
    \label{eq:retrieval}
\end{equation}
where $d$ is a triplet embedding function and $s$ is a dialogue context embedding function.
For implementing $s$, we can use any off-the-shelf PLMs, but for $d$, we need another effective approach that captures the property of the graph.
Therefore, we utilize the Graph Neural Networks (GNNs) for the triplet embedding function $d$ to consider the relational structure between entities in the KG.

More specifically, we consider a set of triplets associated to the entities that appear in the given dialogue context: $\set{(\tte, \ttr, \tte_t) \: \text{or} \: (\tte_h, \ttr, \tte)}{ q(\tte) \subseteq \vx}$, as the retrieval candidates.
Then, to effectively represent triplets consisting of entities and their relations as items, we use GNNs described in Section~\ref{sec:gnn}.
In our triplet retriever, utilizing both nodes and edges, which are equally essential components for the multi-relational graph, is worthwhile to represent an entire triplet.
We adopt the existing node~\citep{GCN} and edge message passing frameworks~\citep{EHGNN}. 
Formally, our triplet embedding function is denoted as follows:
\begin{equation}
\label{eqn:triplet}
\begin{aligned}
d(z;\gG) = \texttt{MLP}([\ve_h \mathbin\Vert \vr \mathbin\Vert \ve_t]), \ 
\ve_h = \texttt{GNN}(\ve^0_h;\gG), \\ 
\vr = \texttt{GNN}(\vr^0;\gG^*), \
\ve_t = \texttt{GNN}(\ve^0_t;\gG),
\end{aligned}
\end{equation}
where $z = (\tte_h, \ttr, \tte_t)$, $q(\tte) = [\vx_i, \ldots, \vx_j]$, $\boldsymbol{0}$ is a zero vector, and $\mathbin\Vert$ is the concatenation operator.
For node embedding $\ve^0$, we reuse the word embedding from $\texttt{Enc}$, if it exists in $\vx$.
For relation embedding $\vr^0$, we use the trainable relation embedding matrix. 
Please refer to Appendix~\ref{appendix:triplet/encoding} for more details.

\subsection{Invariant Graph Encoding}
\label{sec:encoding}
We now specify graph encoding, which aims to condition the structural graph $\gZ$ along with the text sequence $\vx$ over PLMs to generate $\vy$.
Let $\boldsymbol{\psi}(\vx, \gZ)$ be a graph encoding function.
Then, the simplest way to encode graphs into PLMs is to prepend the tokens of entities and relations to the input $\vx$~\citep{PLUG, UDT-QA}.
Formally, given a text $\vx = [x_1, \ldots, x_N]$ and a graph $\gZ = \{(\tta, \ttr_1, \ttb), (\ttb, \ttr_2, \tta), (\tta, \ttr_1, \ttc)\}$, a na\"ive graph encoding is defined as follows: $\boldsymbol{\psi}(\vx, \gZ) = f([a, r_1, b, b, r_2, a, a, r_1, c, x_1, ..., x_N])$ where $a=q(\tta)$, $r_1=q(\ttr_1)$, and so on. Here $f$ is a token embedding and $q$ is a mapping function defined in Section~\ref{sec:prelim}.
However, it violates two important properties for consistent encoding of a multi-relational graph into PLMs: permutation invariance~\citep{deepsets} and relation-inversion invariance, formalized in Definition~\ref{def:permut} and~\ref{def:inverse_relation} as follows:

\begin{definition}
\textbf{(Permutation Invariance)}
\label{def:permut}
\textit{
For any permutation $\pi \in S_n$, $\boldsymbol{\psi}(\vx, \gZ) = \boldsymbol{\psi}(\vx, \pi \cdot \gZ)$.
}
\end{definition}

\begin{definition}
\textbf{(Relation Inversion Invariance)}
\label{def:inverse_relation}
\textit{
Let $\lnot\texttt{r}$ be an inverse relation to $\texttt{r}$, if $(\texttt{a}, \texttt{r}, \texttt{b}) = (\texttt{b}, \lnot\texttt{r}, \texttt{a}) \: \forall \texttt{a}, \texttt{b} \in \mathcal{E}$. Then, $\boldsymbol{\psi}(\vx, \gZ \cup \{(\texttt{a}, \texttt{r}, \texttt{b})\}) = \boldsymbol{\psi}(\vx, \gZ \cup \{(\texttt{b}, \lnot\texttt{r}, \texttt{a})\})$ for any subgraph $\gZ$.
}
\end{definition}

\paragraph{Invariant Graph Encoding}
To satisfy both properties, we consider two operations on a set of triplets up to the na\"ive encoding.
We first define a $\texttt{SORT}$ operator that returns the same output regardless of the order of input set elements, as follows:
\begin{equation}
    \label{eqn:sort}
    \texttt{SORT}(\pi \cdot \gZ) = \texttt{SORT}(\pi' \cdot \gZ), \ \forall \pi, \pi' \in S_n,
\end{equation}
where $S_n$ is a set of all possible permutations for $n$ elements. We then
define a $\texttt{INV}$ operator that adds the inverse triplet of each triplet in $\gZ$, as follows:
\begin{equation}
\fontsize{10pt}{10pt}\selectfont
    \label{eqn:inv}
    \texttt{INV}(\gZ) = \gZ \cup \set{(\tte_t, \lnot \ttr, \tte_h)}{(\tte_h, \ttr, \tte_t) \in \gZ}.
\fontsize{10pt}{10pt}\selectfont
\end{equation}
Based on them, our graph encoding function, $\boldsymbol{\psi} (\vx, \texttt{SORT}(\texttt{INV}(\gZ)))$, satisfies both invariances.

\paragraph{Invariant and Efficient Graph Encoding}
However, the above encoding is not efficient since it requires $\gO(n)$ space complexity with $n$ triplets. Thus, we newly define $\tilde{\boldsymbol{\psi}}$ that encodes the sorted sequence of only the unique entities, as follows:
\begin{equation*}
    \tilde{\boldsymbol{\psi}}(\vx, \texttt{SORT}(\texttt{ENT}(\gZ))) = f([a,b,c,x_1,\ldots,x_N]),
\end{equation*}
where $\texttt{ENT}(\gZ)$ returns the set of unique
entities in $\gZ$.
This encoding meets both invariance properties but also efficient since it only costs $\gO(k)$, for the $k$ entity where $k < n$.
However, as it does not consider the relational information in $\gZ$, we further perturb the token embeddings of each entity $f(\cdot)$ in PLMs with respect to their graph representations in $\gZ$. Specifically, for each entity $\texttt{a} \in \texttt{ENT}(\gZ)$, we apply a learnable affine transformation~\citep{FiLM} on the token embedding of $\texttt{a}$ as follows:
\begin{equation}
\label{eqn:perturb}
\begin{aligned}
\boldsymbol{\beta}(f(a), \gZ) = (1 + \boldsymbol{\gamma}) * f(a) + \boldsymbol{\delta}, \\
\boldsymbol{\gamma}, \boldsymbol{\delta} = \texttt{MLP}(\boldsymbol{\eta}), \ 
\boldsymbol{\eta} = \text{\texttt{R-GNN}}(f(a);\gZ),
\end{aligned}
\end{equation}
where \texttt{MLP} is a Multi-Layer Perceptron, $\boldsymbol{\beta}:\mathbb{R}^d\rightarrow\mathbb{R}^d$ perturbs the embedding according to $\gZ$, \texttt{R-GNN} is the relation-aware GNN~\citep{CompGCN}.
In sum, we denote a relation-aware and invariant yet efficient encoding $\boldsymbol{\psi}^*$, defined as follows:
\begin{equation*}
    \boldsymbol{\psi}^*(\vx, \gZ) =
    \boldsymbol{\beta}(
    \tilde{\boldsymbol{\psi}}(
    \vx, \texttt{SORT}(\texttt{ENT}(\mathcal{Z}))), 
    \texttt{INV}(\gZ)
    ).
\end{equation*}
We conclude that our graph encoding satisfies both properties. 
For further details on the proof and comprehensive illustration, please refer to Appendix~\ref{appendix:proof}.

\begin{figure*}
    \centering
    \includegraphics[width=0.99\linewidth]{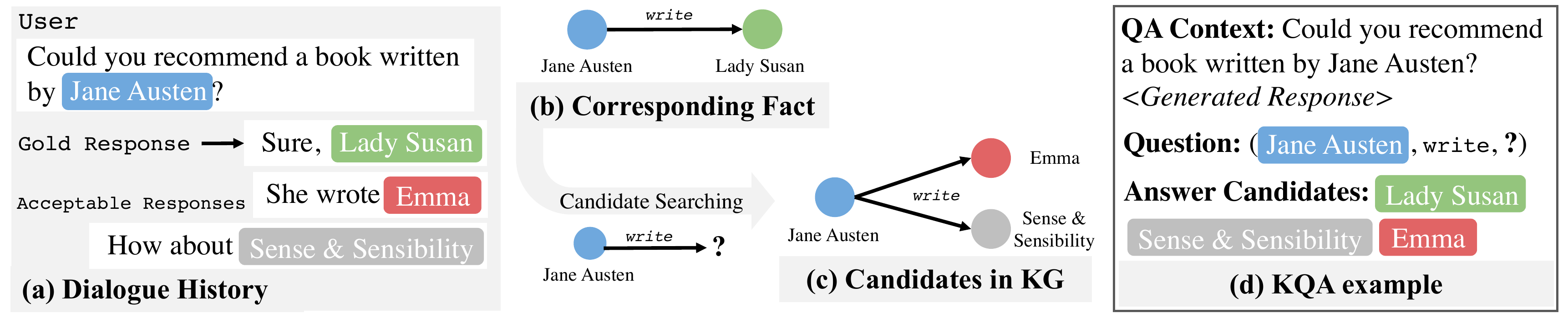}
    \caption{\small \textbf{KQA.} (Left) An example where multiple responses are acceptable. (Middle) In KG, we search for the fact that reflects the relation between entities within the user input and gold response (b), and then search candidate facts from the KG (c). (Right) Corresponding KQA example. If a generated response contains the one of answer candidates, the KQA can predict it (success). 
    }
    \vspace{-0.15in}
    \label{fig:kqa}
\end{figure*}

\subsection{Knowledge-Grounded Generation with Graph-Text Contrastive Learning}
\label{sec:contrastive}
Our framework now can retrieve and encode the context-relevant subgraph given the user input.
Then, reflecting the subgraph into the model is important when generating a knowledge-grounded response.
The generative model should be able to generate different sequences when providing different subgraphs, for the same dialogue history. 

However, we only access the single ground-truth response regardless of the retrieved knowledge, while the generative model is trained with a teacher forcing. Thus, this setting can raise the problem of \textit{exposure bias}~\citep{exposurebias}: the model is never exposed to other generated tokens during training. 
To overcome such limitations, we introduce a graph-text contrastive learning method. 
Formally, for a single pair of a graph and text, the contrastive learning objective is defined as follows:
\begin{equation}
\label{eqn:cont}
    \gL_{cont} = \frac{1}{2} \log \frac{\exp(\texttt{sim}(\zeta(\vz), \xi(\vh)) / \tau)}
    {\sum_{\vh'} \exp(\texttt{sim} (\zeta(\vz), \xi(\vh')) / \tau) } + 
    \frac{1}{2} \log \frac{\exp(\texttt{sim}(\zeta(\vz), \xi(\vh)) / \tau)}
    {\sum_{\vz'} \exp(\texttt{sim} (\zeta(\vz'), \xi(\vh)) / \tau)}, 
\end{equation}
where $\vz = \frac{1}{m} \sum_{i=1}^{m} \tilde{\vz}_i$ is the average representations of the graph from $\texttt{Enc}(\boldsymbol{\psi}^*(\vx, \gZ)) = [\tilde{\vz}_1, \ldots, \tilde{\vz}_{m}, \vz_1, \ldots, \vz_N]$, $\vh = \frac{1}{T}\sum_{t=1}^T \vh_t$ is the mean of decoder representations, \texttt{sim} is the cosine similarity, $\zeta$ and $\xi$ are learnable linear projection layers, and $\tau$ is a learnable temperature parameter. Furthermore, $\sum_{\vh'}$ and $\sum_{\vz'}$ indicate the summation over negative samples, which are other texts or graphs within a same mini-batch.

\subsection{Training}
\label{sec:training}
We train the entire model by maximizing the log-likelihood $\log p(\vy|\vx)$ defined in~\autoref{eqn:likelihood} with respect to parameters of both the retriever $\phi$ and the generator $\theta$. 
Since computing the marginal probability over entire subgraphs is infeasible, we approximate it by summing over $k$ sampled subgraphs~\citep{REALM, RAG}.
Our end-to-end training objective for retrieval-augmented generation is then defined as follows:
\begin{equation}
    \label{eqn:ret}
    \mathcal{L}_{ret} = \log \sum_{\gZ \subseteq \Pi} p_\phi (\gZ | \vx) p_\theta (\vy|\vx,\gZ),
\end{equation}
where $\Pi=\text{\texttt{samplek}}(p_\phi(\cdot | \vx))$ denotes sampling $k$ subgraphs over the subgraph distribution and each subgraph sampling is decomposed into sampling $n$ triplets from $p_\phi(z_i|\vx) \forall i\in [1,n]$ as in Section~\ref{sec:retriever}.
We further assume that the gold subgraph $\gZ^*$ is partially available in training. 
In this case, we utilize the following supervised loss to train the retriever: $\mathcal{L}_{sup} = \log p_\phi(\gZ^* | \vx)$.
By combining all objectives in~\autoref{eqn:cont},~\ref{eqn:ret}, and $\mathcal{L}_{sup}$, our training objective is defined as $\gL = \gL_{ret} + \gL_{sup} + \gL_{cont}$.

\section{KQA Metric: Knowledge-verifying QA}
\label{sec:kqa}
Existing automatic evaluation metrics, namely BLEU and ROUGE~\citep{BLEU, ROUGE}, are limited in that they only consider the lexical overlaps of words without measuring the factual correctness. For instance, as shown in~\autoref{fig:kqa} (a), there could be multiple correct responses, but existing metrics score them lower due to the lexical mismatch with the gold response. To solve this issue, we propose \textbf{K}nowledge-verifying \textbf{Q}uestion \textbf{A}nswering (\textbf{KQA}) which measures whether generated responses contain factually correct knowledge given the dialogue history.
To realize this, we formulate extractive QA task~\citep{SQuAD} by automatically deriving QA pairs from the dialogue and the large-scale KG in each dataset (See~\autoref{fig:kqa}).
Then, we fine-tune BERT~\citep{BERT} on synthetic KQA pairs to build a QA model.
To evaluate generated responses, we concatenate the dialogue history and the generated response then forward it into the trained QA model.
If the QA model yields the correct answer, we regard this case as the generated response containing accurate knowledge.
See Appendix~\ref{appendix:supple:implementation} for details.
\begin{table*}
	\centering
	\caption{\small Experimental results on the OpendialKG dataset with the T5-small model. $\dagger$ indicates the model under the oracle setting, which uses the gold facts even in the test time. The best results are emphasized in bold.}
	\resizebox{0.99\textwidth}{!}{
    \renewcommand{\arraystretch}{0.85}
	\begin{tabular}{clcccccccccc}
		\toprule
    		              & & \multicolumn{2}{c|}{\textbf{KQA}} & \multicolumn{4}{c|}{\textbf{BLEU}} & \multicolumn{3}{c|}{\textbf{ROUGE}} & \multicolumn{1}{c}{\textbf{Unigram}}  \\
		& {\textbf{Method}} & EM & F1 & B-1 & B-2 & B-3 & B-4 & R-1& R-2& R-L & F1  \\
		\midrule[0.8pt]
		\multirow{5}{*}{\textit{Baselines}} 
		& \textbf{No Knowledge} &  12.25 & 20.69 & 15.79 & 9.19 & 5.61 & 3.43 & 19.67 & 7.13 & 19.02 & 22.21 \\
		& \textbf{All Knowledge} & 43.58 & 50.60 & 15.95 & 9.98 & 6.72 & 4.65 & 20.96 & 8.50 & 20.21 & 24.34 \\
		& \textbf{Space Efficient} \textit{(series)} & 36.60 & 42.64 & 16.15 & 10.03 & 6.66 & 4.50 & 21.15 & 8.56 & 20.44 & 24.55 \\
		& \textbf{Space Efficient} \textit{(parallel)}& 38.54 & 44.34 & 16.33 & 10.22 & 6.81 & 4.64 & 21.42 & 8.85 & 20.68 & 24.87\\
		& \textbf{EARL} & 32.47 & 35.88 & 11.49 & 6.34 & 4.06 & 2.75 & 15.36 & 4.37 & 14.61 & 16.88 \\
		& \textbf{DiffKG} & 12.25 & 20.99 & 15.68 & 9.13 & 5.60 & 3.46 & 19.50 & 7.07 & 18.84 & 22.26 \\
		\midrule[0.5pt]
		\multirow{4}{*}{\textit{\shortstack{Retrieval\\variants}}} 
		& \textbf{Random Retrieval} & 31.72 & 38.95 & 15.70 & 9.52 & 6.12 & 3.99 & 20.21 & 7.88 & 19.55 & 23.28 \\
		& \textbf{Sparse Retrieval} (BM25) & 29.50 & 36.96 & 15.63 & 9.44 & 6.05 & 3.96 & 20.05 & 7.67 & 19.37 & 23.10 \\
		& \textbf{Dense Retrieval} (Bi-encoder) & 46.17 & 52.52 & 16.67 & 10.44 & 7.05 & 4.91 & 20.41 & 8.38 & 19.66 & 23.85 \\
		& \textbf{Dense Retrieval} (Poly-encoder) & 46.05 & 52.57 & 17.56 & 11.01 & 7.45 & 5.18 & 20.66 & 8.46 & 19.87 & 24.24   \\
		\midrule[0.5pt]
		\multirow{3}{*}{\textit{Ours}}
		& \textbf{SURGE} \textit{(unsupervised)}& 48.49 & 55.77 & \bf 17.77 & \bf 11.30 & \bf 7.69 & \bf 5.36 & \bf 21.64 & \bf 9.14 & \bf 20.75 & \bf 25.24 \\
		& \textbf{SURGE} \textit{(semi-supervised)}  & \textbf{51.00} & 57.63 & 17.70 & 11.21 & 7.61 & 5.28 & 21.43 & 8.85 & 20.57 & 25.07 \\
		& \textbf{SURGE} \textit{(contrastive)} &
	    50.45 &	\textbf{57.70} & 17.29 & 11.04 & 7.54 & 5.28 &	21.35 & 8.98 & 20.48 & 25.10 \\
		
		\midrule[0.5pt]
		\multirow{2}{*}{\textit{Oracle}}
		& \textbf{Gold Knowledge}$^\dagger$ &  63.32 & 67.90 & 18.47 & 12.79 & 9.32 & 6.92 & 24.93 & 11.97 & 24.03 & 28.82 \\
		& \textbf{Gold Response} & 93.30 & 95.21 & 100.0 & 100.0 & 100.0 & 100.0 & 100.0 & 100.0 & 100.0 & 100.0 \\
		\bottomrule
	\end{tabular}
	}
	\label{main_exp}
	\vspace{-0.125in}
\end{table*}
\begin{figure*}
    \begin{minipage}{0.38\textwidth}
        \centering
        \captionsetup{type=table}
        \captionof{table}{\small Experimental results on the KOMODIS dataset with the T5-small model. For full experimental results, please see Supplementary Files.}
        \resizebox{1.0\textwidth}{!}{
        \renewcommand{\arraystretch}{0.8}
        \renewcommand{\tabcolsep}{1.0mm}
        \begin{tabular}{lccccc}
        \toprule
         & \multicolumn{2}{c}{\bf KQA} & \multicolumn{3}{c}{\bf BLEU}  \\
         \cmidrule(l{2pt}r{2pt}){2-3} \cmidrule(l{2pt}r{2pt}){4-6}
         & {EM} & {F1} & {B-1} & {B-2} & {F1} \\
        \midrule
        \textbf{Random} & 12.41 & 14.17 & 7.74 & 4.02 & 16.29 \\
        \textbf{SE} \textit{(series)} & 12.41 & 14.70 & 8.34 & 5.13 & 17.37 \\
        \textbf{SE} \textit{(parallel)} & 16.46 & 18.70  & 9.33 & 5.66 & 17.72 \\
        \midrule
        \textbf{SURGE} (Ours) & \bf 17.30 & \bf 19.50  & \bf 11.51 & \bf 7.18 & \bf 19.51 \\
        \bottomrule
        \end{tabular}
        \label{tab:komodis}
        }
    \end{minipage}
    \hfill
    \begin{minipage}{0.38\textwidth}
        \centering
        \captionsetup{type=table}
        \captionof{table}{\small Knowledge-grounded generation results by using the modified gold subgraphs instead of the retrieved ones, to evaluate the efficacy of contrastive learning, with F1 and KF1 as metrics.}
        \resizebox{0.90\textwidth}{!}{
            \renewcommand{\arraystretch}{1.15}
    	    \begin{tabular}{lcc}
    		\toprule
    		{\textbf{Method}}  & F1 & KF1 \\
    		\midrule[0.8pt]
    		 \textbf{SURGE} (unsupervised) & 27.78 & 24.09 \\
    		 \textbf{SURGE} (semi-supervised) & \textbf{28.30} & 26.38\\
    		 \textbf{SURGE} (contrastive) & 28.17 & \textbf{27.58} \\
     		\bottomrule
    	    \end{tabular}
    	}
    	\label{consitent}
    \end{minipage}
    \hfill
    \begin{minipage}{0.2\textwidth}
        \centering
        \includegraphics[width=1\linewidth]{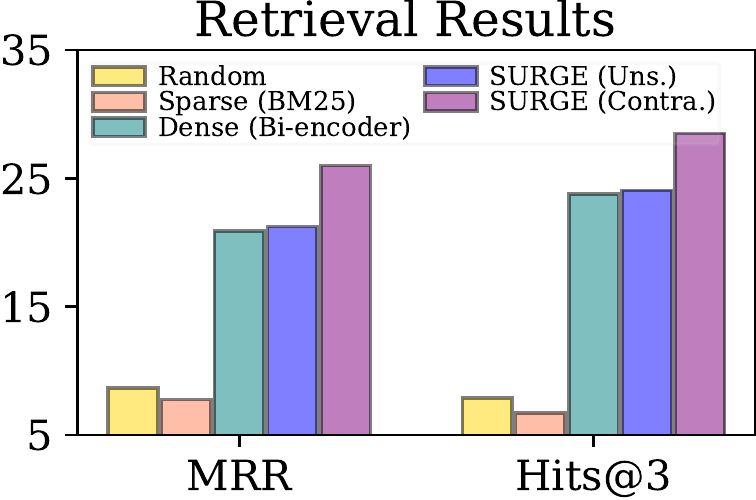}
    	\caption{\small Fact retrieval results on the OpendialKG dataset, with the metrics of MRR and Hits@3.}
    	\label{retrieval_exp}
    \end{minipage}
    \vspace{-0.175in}
\end{figure*}

\section{Experiment}
\label{sec:experiment}
\subsection{Experimental Setup}
We conduct experiments on \textbf{OpendialKG}~\citep{opendialkg}, which consists of 15K dialogues with 91K utterances associated with a large-scale KG, namely Freebase~\citep{Freebase} with 100k entities and 1M facts. We note that, among them, $49\%$ of the utterances come with gold knowledge, whereas others are not. We randomly split the dataset into training (70\%), validation (15\%), and test sets (15\%).
We also use \textbf{KOMODIS}~\citep{KOMODIS}, which contains 7.5K dialogues associated with the KG having 88k facts.
As retrieval candidates, we use 1-hop KG associated with entities in the given dialogue for OpendialKG and 2-hop KG for KOMODIS.
Except~\autoref{tab:komodis}, most of the experiments are conducted on OpendialKG.
We use \textbf{T5-small}~\citep{T5} for all experiments.
For more details, see Appendix~\ref{appendix:setup}.

\subsection{Baselines and Our Models}
\textbf{No Knowledge.} The model only with the dialogue history.
\textbf{All Knowledge.} The model with entire facts within a $k$-hop subgraph. 
\textbf{Gold Knowledge.} The model with the exact gold knowledge if it exists.
\textbf{Space Efficient Encoding.} The model with all facts from the $k$-hop subgraph. We use two variants from~\citep{efficientdialogue}, namely \textit{series} and \textit{parallel}.
\textbf{EARL.} The latest RNN-based model, where the entities are conditioned in response generation~\citep{EARL}.
\textbf{DiffKG.} PLM-based model with differentiable path traversal~\citep{DiffKG}.
\textbf{Random/Sparse Retrieval.} The model with selected facts via random sampling or sparse retrieval~\citep{BM25}.
\textbf{Dense Retrieval.} A variant of our framework where T5 encoder~\citep{T5} is used for $d$ in ~\autoref{eqn:triplet} instead of GNNs similar to Bi- and Poly-encoder~\citep{Polyencoder}.
\textbf{SURGE (unsupervised).} Ours with retrieved context-relevant facts from $k$-hop subgraph, where the retrieval is trained without any supervision.
\textbf{SURGE (semi-supervised).} Ours but the retriever is trained with $\mathcal{L}_{sup}$ if the gold  exists.
\textbf{SURGE (contrastive).} Ours with $\mathcal{L}_{ret}$, $\mathcal{L}_{sup}$, and $\mathcal{L}_{cont}$ in Section~\ref{sec:training}.

\subsection{Evaluation Metrics}
We use BLEU~\citep{BLEU}, ROUGE~\citep{ROUGE}, and F1 score as metrics. We also use our new metric, KQA (\S~\ref{sec:kqa}), which measures whether the generated responses contain proper knowledge.
In~\autoref{consitent}, we use Knowledge F1 (KF1)~\citep{RAG-FiD} to measure unigram overlaps between the retrieved knowledge and generated response.

\subsection{Experimental Results and Analysis}
In~\autoref{main_exp}, we report the knowledge-grounded response generation performances of baselines and our SURGE on OpendialKG. As shown in~\autoref{main_exp}, our models significantly outperform all the baseline models, excluding oracles, in all evaluation metrics. The high BLEU, ROUGE, and F1 refer that ours sufficiently learns the syntactic and semantic structure of the responses. Our models also achieve high F1 and EM scores in KQA. 
The high KQA scores indicate that the generated responses are formed with the correct facts, which are relevant to the dialog context. 
Even the baseline models such as \textit{All Knowledge}, \textit{Space Efficient Encoding}~\citep{efficientdialogue}, \textit{EARL}~\citep{EARL}, and \textit{DiffKG}~\citep{DiffKG}, which are provided with all of $k$-hop facts, underperform than ours.
The result demonstrates that selecting relevant knowledge is critical in knowledge-augmented response generation.
In~\autoref{tab:komodis}, we additionally report the experimental results on KOMODIS to show the applicability of our method to other datasets. Our SURGE (\textit{contrastive}) also outperforms other baselines in KOMODIS. 
For results with all metrics, please see Appendix~\ref{appendix:exp:komodis}.
\begin{table*}[t!]
    \centering
    \centering
    \caption{\small Examples of responses from the baseline (Space Efficient, parallel) and responses from SURGE.}
        \resizebox{0.97\textwidth}{!}{
        \renewcommand{\arraystretch}{0.99}
        \begin{tabular}{p{0.05in}p{1.5in}p{1.5in}p{2.0in}p{2.5in}}
        \toprule
         & Context  & Gold response & Baseline response & SURGE response \\
        \midrule[0.8pt]
        \textcolor{red}{(a)} 
        & I loved Moby Dick. Can you recommend something similar?
        & It was written by Herman Melville in 1851. It's sometimes called The Whale.
        & Moby Dick is a sailor. Do you like her work?
        & Moby Dick was written by Herman Melville. He also wrote The Whale.
        \\
        \textcolor{blue}{(b)} 
        & Do you know anything the actor Adam Brown?
        & Yes, he was in the movie The Hobbit: An Unexpected Journey.
        & Adam Brown starred in King Kong. Have you seen it?
        & Adam Brown starred in The Hobbit: The Desolation of Smaug and The Hobbit: The Battle of the Five Armies. \\
        \midrule[0.5pt]
       \multicolumn{3}{l}{\textcolor{red}{(a) Retrieved Subgraph from SURGE}}& 
       \multicolumn{2}{l}{\textcolor{blue}{(b)Retrieved Subgraph from SURGE}} \\
       \multicolumn{3}{l}{(Moby Dick; The Whale, written\_by, Herman Melville)}& 
       \multicolumn{2}{l}{(The Hobbit: The Battle of the Five Armies, starred\_actors, Adam Brown)} \\
     \multicolumn{3}{l}{(Moby Dick, written\_by, Norman Corwin)}& 
       \multicolumn{2}{l}{(The Hobbit: An Unexpected Journey, starred\_actors, Adam Brown)} \\
     \multicolumn{3}{l}{(Moby Dick, written\_by, Ray Bradbury))}& 
       \multicolumn{2}{l}{(The Hobbit: The Desolation of Smaug, starred\_actors, Adam Brown)} \\
       \bottomrule
       \end{tabular}
    }
    \label{tab:example}
\end{table*}
\begin{figure*}
\begin{minipage}[t]{0.7\textwidth}
\vspace{0pt}
    \centering
    \captionsetup{type=table}
    \captionof{table}{\small Experimental results on OpendialKG with additional three metrics other than KQA for measuring whether the generated responses contain appropriate knowledge.}
    \resizebox{0.99\textwidth}{!}{
    \renewcommand{\arraystretch}{0.99}
\begin{tabular}{lcccccccccc}
    \toprule
                  & \multicolumn{2}{c|}{\textbf{KQA}} & \textbf{Knowledge}  &  \textbf{Entity} &  \textbf{String} \\
    {\textbf{Method}} & EM & F1 & \textbf{F1} & \textbf{F1} & \textbf{Matching} \\
    \midrule[0.8pt]
        \textbf{All Knowledge} & 43.58 & 50.60 & 18.91 & 21.10 & 44.25 \\
    \textbf{Space Efficient} \textit{(Parallel)} & 38.54 & 44.34 & 17.43 & 18.93 & 40.56 \\
        \textbf{Dense Retrieval} \textit{(Poly-encoder)} & 46.05 & 52.57 & 19.72 & 21.46 & 48.41 \\
    \midrule
    \textbf{SURGE} \textit{(ours, semi-supervised)}  & \bf 51.00 & \bf 57.63 & \bf 21.87 & \bf 23.03 & \bf 55.79 \\
        \midrule
    \textbf{Gold Response} \textit{(oracle)}  & 93.30 & 95.21 & 28.62 & 29.06 & 85.75 \\
    \bottomrule
\end{tabular}
}
\label{tab:knowledge}
\end{minipage}%
\begin{minipage}[t]{0.31\textwidth}
\vspace{0pt}
\centering
\includegraphics[width=0.9\textwidth]{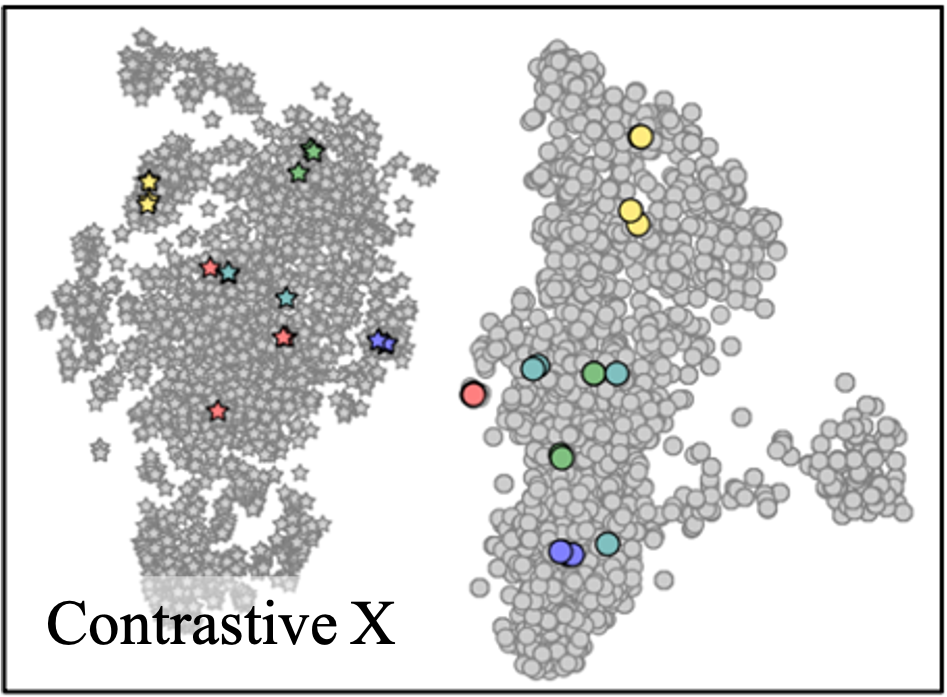}
\includegraphics[width=0.9\textwidth]{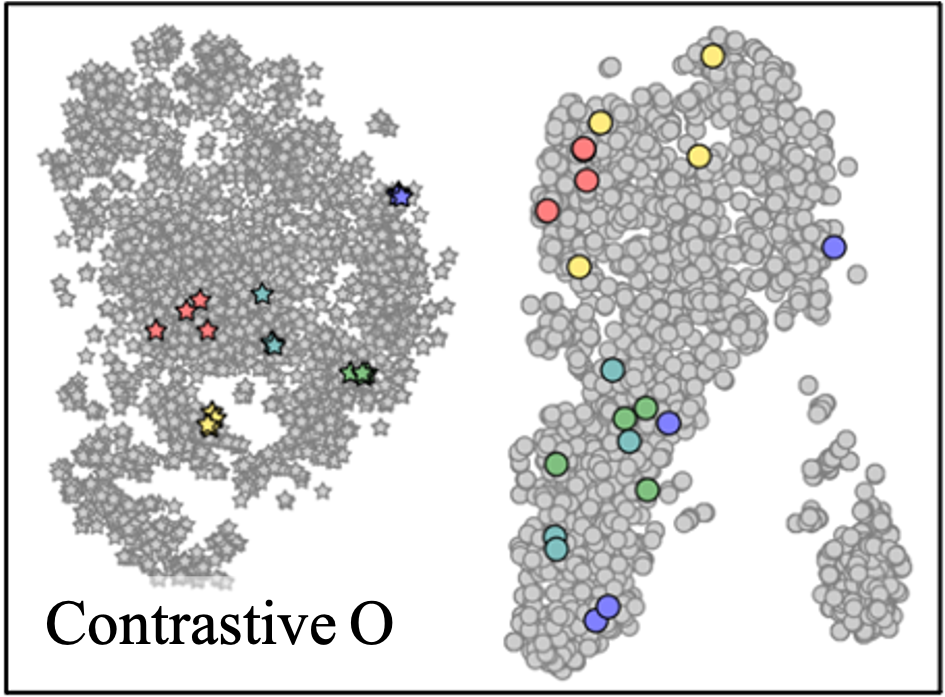}
\caption{\small Embedding visualization of graph (star) - text (circle) contrastive learning.}
\label{fig:contrastive}
\end{minipage}
\hfill
\begin{minipage}{0.69\textwidth}
\vspace{-83pt}
\begin{minipage}[t]{0.5\textwidth}

\centering
\captionsetup{type=table}
\captionof{table}{\small Performance comparisons of variants of graph encodings, described in Section~\ref{sec:encoding}. (Inv.= Invariant)}
\resizebox{1.0\textwidth}{!}{
\renewcommand{\arraystretch}{1.0}
\begin{tabular}{lcc}
    \toprule
    &  \textbf{KQA} & \textbf{Knowledge} \\
    {\textbf{Method}} & {\textbf{F1}}  & \textbf{Length} \\
    \midrule[0.8pt]
    \textbf{Na\"ive Encoding}& 55.23  & 68.21 \\ 
    \textbf{Invariant}      & 56.78  & 121.82 \\  
        \textbf{Inv. \& Efficient} (entity only) & 49.99  & \bf 15.42 \\
    \textbf{Inv. \& Efficient} (w/~\autoref{eqn:perturb}) & \bf{57.63} & \textbf{15.42} \\ 
    \bottomrule
\end{tabular}
}

\label{graph_encoding_exp}
\end{minipage}
\hfill
\begin{minipage}[t]{0.48\textwidth}
\centering
\captionsetup{type=table}
\captionof{table}{\small Human evaluation on \textbf{Consis}tency, \textbf{Info}rmativeness, and \textbf{Fluency} with bold marks $p < 0.05$.}
\resizebox{0.99\textwidth}{!}{
\renewcommand{\arraystretch}{0.9999}
\begin{tabular}{lccc}
    \toprule
   {\textbf{Method}} & Consis. & Info. & Fluency \\
    \midrule[0.8pt]
    \textbf{All Knowledge} &\bf 2.52 & 1.99 & 2.62 \\
    \textbf{Space Efficient} & 2.47 & 1.75 & 2.46 \\
    \midrule[0.5pt]
    \textbf{SURGE} \textit{(ours)} & \textbf{2.71} & \textbf{2.39} & \textbf{2.92} \\
    \bottomrule
\end{tabular}
}

\label{tab:human_eval}
\end{minipage}
\end{minipage}

\end{figure*}

\paragraph{Knowledge Retrieval}
\autoref{retrieval_exp} shows performances of retrievers, for which we measure the performance on 45\% of test dialogues containing the gold knowledge, with Mean Reciprocal Rank (MRR) and Hits@k as metrics.
Our models outperform all baselines. Further, our model with contrastive learning and semi-supervised retriever training outperforms an unsupervised version. 
See Appendix~\ref{appendix:examples} for examples.

\paragraph{Knowledge-Grounded Generation} 
We conduct an ablation study on our models to validate the knowledge consistency performance of the response generation by computing the Knowledge F1 (KF1) score~\citep{RAG-FiD}. 
We use the gold knowledge rather than the retrieved one to focus solely on the case where a given knowledge is consistently reflected in the generated responses.
We randomly modify the tail entity of each gold knowledge to ensure that responses are generated from the given knowledge rather than the trained knowledge.
\autoref{consitent} shows that our model with a contrastive learning term outperforms all others in the KF1, implying that the generated responses accurately reflect the encoded knowledge. 

\paragraph{Retrieval and Generation Examples}
\autoref{tab:example} shows the examples of generated responses along with the retrieved knowledge.
We compare our SURGE against \textit{Space Efficient (parallel)} baseline.
In example (a), the baseline response contains an incorrect fact distracted by the contextually irrelevant entity `sailor'.
Contrarily, SURGE successfully retrieves relevant facts from the KG and then generates the factually correct response.
In (b), the baseline generates the response with a wrong fact, meanwhile, SURGE retrieves context-relevant facts and generates an informative response.

\paragraph{Automatic Evaluations on Knowledge Groundedness}
In~\autoref{tab:knowledge}, we measure Knowledge F1 (KF1 in~\autoref{consitent}), string matching (check whether at least one of answer entities exists the generated response), and entity F1 (measuring F1 score with each entity in answer candidates) for representative baselines and our SURGE (semi-supervised) in OpendialKG. For KF1, we measure the F1 score regarding the concatenation of the question (head entity and relation) and all answer candidates (available tail entities) in KQA as the gold response. The results show that all metrics show the same tendency with KQA and our proposed method still outperforms other baselines by generating responses with more proper knowledge.
See Appendix~\ref{appendix:exp:groundedness} for more details.

\paragraph{Sensitive Analysis on Graph Encoding}
We conduct an analysis on graph encoding variants introduced in Section~\ref{sec:encoding}. The knowledge length in~\autoref{graph_encoding_exp} indicates the average token length used for graph encoding. Our encoding $\boldsymbol{\psi}^*$ with ~\autoref{eqn:perturb} performs the best against other variants while using the lesser space at the graph encoding phase. 

\paragraph{Human Evaluation}
We sample 30 responses of SURGE, \textit{All Knowledge}, and \textit{Space Efficient} on the test set of OpendialKG, then conduct a human study of them. We recruit $46$ annotators and ask them to evaluate the quality of the generated responses with consistency, informativeness, and fluency criteria using a 3-point Likert-like scale. As shown in~\autoref{tab:human_eval}, ours obtains significantly higher scores than others in all criteria, which is another evidence that our framework generates consistent, informative, and fluent responses.
We observe that the informativeness score and KQA F1 score have a 0.42 Pearson correlation coefficient. This allows us to confirm that our KQA metric positively correlates with the human evaluation results.

\paragraph{Embedding Space Visualization}
We further visualize the latent space of graph and text learned from~\autoref{eqn:cont} in~\autoref{fig:contrastive}. The visualization shows that, for the same dialogue with different subgraphs, our SURGE with graph-text contrastive learning (right) generates distinct response embeddings pertaining to different subgraphs, unlike the one without contrastive learning which shows less variety over responses for the same dialogue (left). We include zoomed~\autoref{fig:contrastive} in Appendix~\ref{appendix:examples}.
\section{Conclusion}
In this work, we proposed a novel end-to-end framework for knowledge-grounded dialogue generation which retrieves context-relevant subgraph, encodes a subgraph with the text, and generates natural and informative responses based on the retrieved subgraph, called as \textbf{SU}bgraph \textbf{R}etrieval-augmented \textbf{GE}neration (\textbf{SURGE}). 
Our results demonstrate the effectiveness of our framework in both quantitative and qualitative experiments in knowledge retrieval and response generation tasks. The analysis shows the contribution of each proposed component: retrieval, encoding, and graph-text representation learning. Our work suggests a new direction to generate informative responses for knowledge graph-based dialogue task by empirically showing the importance of retrieving the more relevant subgraph knowledge rather than using all the relevant knowledge graphs when generating knowledge-grounded responses.


\bibliography{reference}


\clearpage
\appendix
\section*{Appendix}

\section{Limitations}
As discussed in~\autoref{appendix:examples}, our work is limited in a variety of dimensions primarily in terms of the lack of a well-formulated public dataset, retrieval accuracy, and generation quality.
First, the public benchmark dataset for knowledge-consistent dialogue generation is highly limited. Despite the fact that there are several public Knowledge Graphs (KGs) available~\citep{wikidata,Freebase}, only one dataset~\citep{opendialkg} provides the diverse set of dialogue and its corresponding large-scale KG.
This circumstance may limit the rigorous evaluation of our framework's adaptability in various settings.
Future work may study applying our approach for a wider range of dialogue datasets based on Wikipedia~\citep{WoW} by leveraging existing public large-scale KG such as Wikidata~\citep{wikidata}.
Second, the search space for retrieving context-relevant subgraphs can be expanded. Our SURGE framework now runs on a $k$-hop KG that is rooted in entities in the given dialogue history. Finding the entity within the text, on the other hand, necessitates precise named entity extraction and entity linking.
Therefore, future work may investigate extending our approach to a framework that can retrieve the context-relevant subgraph among the entire KG instead of $k$-hop KG.
Third, there is still room for improvement in generation quality since we generate knowledge-enhanced responses with a small-scale Pre-trained Language Model (PLM) for efficiency. Such PLMs occasionally fail to generate high-quality natural sentences~\citep{T5}. Future work could improve generation quality based on a larger PLM.

\section{Broader Impacts}
Our proposed knowledge-grounded dialogue generation model is essential for designing user-friendly real-world AI systems. 
Among various types of dialogue generation models, knowledge-grounded dialogue models are trained to interact with users and convey factual information to users in natural languages.
Their conversational features can be adapted to any user interface that connects the bilateral interaction between humans and computers.
We believe that conversational interfaces can enhance the users' experiences and reduce the users' efforts in learning how to use the systems. 
However, knowledge-grounded dialogue models can become vulnerable to generating offensive and harmful content or responses with misinformation depending on the users or data. When deploying the models in the real world, in addition to generating realistic responses, they also need to be robust to adversarial feedback from malicious users and biases inherited in pre-training or training corpus, or else they could malfunction. Therefore, along with the quantitative and qualitative evaluations of generated responses, it would be worthwhile to examine the robustness of the dialogue models.

\section{Notations}
We organize the notations we used for formally describing our method in~\autoref{table:notation}.
\begin{table*}[htbp]\caption{A list of notations that we used for defining our method.}
\centering 
\begin{tabular}{r c p{10cm} }
\toprule
$\gV$ &  & pre-defined vocabulary of tokens for pre-trained language models (text) \\
$\gE$ &  & pre-defined vocabulary of entities (symbol) \\
$\gR$ & & pre-defined vocabulary of relations (symbol) \\
$\tta, \ldots \ttz$ & & knowledge graph symbols written in typewrite font \\
$\vx$ & & input sequence (vector) \\
$x_1, \ldots, x_N$ && input tokens (scalar) \\
$\vy = [y_1, \ldots, y_T]$ && output sequence and tokens \\
$\gG$  && multi-relational graph, such as knowledge graph \\
$\gZ$  && retrieved subgraph: $\gZ \subset \gG$ \\ 
$z$   && triplet (edge): $z \in \gZ$ \\
$q$  && tokenization (mapping) function of KG symbol to the sequence of tokens\\
$s(\cdot)$ && text representation function for retrieval \\
$d(\cdot)$ && triplet representation function for retrieval \\
\texttt{Enc} && Transformer Encoder \\ 
\texttt{Dec} && Transformer Decoder\\ 
$f$ && token (word) embedding function\\ 
$\theta$ && generator parameter\\ 
$\phi$ && retriever parameter \\ 
$\boldsymbol{\psi}$ && set encoding function \\
$\boldsymbol{\beta}$ && perturbation function \\
$\pi$ && set permutation \\
$n$   && the number of triplets in a retrieved subgraph $\gZ$\\
$k$   && the number of samples in a marginalization term\\
$\vz$ && encoder hidden state (single token) \\
$\mZ$ && encoder hidden states (sequence of tokens) \\
$\vh$ && decoder hidden state (single token) \\
$\mH$ && decoder hidden states (sequence of tokens) \\
$\mX$ && input embeddings after token embedding function (sequence) \\
$\mY$ && output embeddings after token embedding function (sequence) \\
\bottomrule
\end{tabular}
\label{table:notation}
\end{table*}

\section{Intuitions \& Proofs for Graph Encoding}
\label{appendix:proof}

\subsection{Intuitions}
In Section 3.4, we focus to introduce the novel graph encoding which meets both permutation and relation inversion invariances. However, one may draw the question of why such invariances are important for graph encoding with the pre-trained language models (PLMs) and need more detailed explanations on this.

First of all, we want to recapitulate why the permutation invariance is important in encoding multi-relational graphs into the PLMs along with the text sequence. As noted in Section 3.1, PLMs are permutation sensitive since the meaning of the sentence can vary when we change the order of words in the sentence (e.g., "A is born in C" $\ne$ "C is born in A"). However, the multi-relational graphs are permutation invariant since they are represented as a set of triplets. For instance, given the multi-relational graphs with two triplets, \{($\tta$, $\texttt{born-in}$, $\ttc$), ($\ttb$, $\texttt{born-in}$, $\ttd$)\}, the order of elements (triplets) does not affect the entire semantic of the graph. (e.g.,  \{($\tta$, $\texttt{born-in}$, $\ttc$), ($\ttb$, $\texttt{born-in}$, $\ttd$)\} $=$ \{($\ttb$, $\texttt{born-in}$, $\ttd$), ($\tta$, $\texttt{born-in}$, $\ttc$)\}).

With a na\"ive encoding, the PLM yields different representations for different orders of triplets in the subgraph. Therefore, if the PLM is only fine-tuned with the input [A, born-in, C, B, born-in, D, where was A born?], there is no guarantee that the PLM will output the exact same response given the input with a permuted subgraph [B, born-in, D, A, born-in, C, where was A born?] in the inference since the PLM is order-sensitive due to its positional encoding. In order to prevent the aforementioned scenarios, we decide to design the permutation-invariant graph encoding which yields stable results regardless of the order of triplets in the graph.
Similarly, the inversion of the triplet yields the same semantic (e.g., \{($\tta$, $\texttt{born-in}$, $\ttc$)\} = \{($\ttc$, $\lnot\texttt{born-in}$, $\tta$)\}), but the graph encoding without considerations for the inverse relation results in different representations from PLM given the triplet and its inversed one. 

\subsection{Proofs}
In this section, we first show that a na\"ive encoding function $\boldsymbol{\psi}$ in Section~3.4 is neither permutation invariant nor relation inversion invariant, formalized in Proposition~\ref{prop:naive}. After that, we prove that our invariant and efficient encoding function $\boldsymbol{\psi}^*$ with graph-conditioned token embedding perturbation is both permutation invariant and relation inversion invariant, formalized in Proposition~\ref{prop:ours}.

\begin{proposition}
\label{prop:naive}
A na\"ive encoding function $\boldsymbol{\psi}$ is neither permutation invariant nor relation inversion invariant.
\end{proposition}
\vspace{-0.2cm}
\begin{proof}
We prove this by contradiction. 

Suppose $\vx = [x_1, \ldots, x_n]$ and $\gZ = \{(\tta, \ttr_1, \ttb), (\ttb, \ttr_2, \tta), (\tta, \ttr_1, \ttc)\}$. Moreover, let $\gZ' = \{(\ttb, \ttr_2, \tta), (\tta, \ttr_1, \ttb), (\tta, \ttr_1, \ttc)\}$ be one of permutations of $\gZ$ with the permutation order $\pi = (2,1,3)$. 

With a na\"ive encoding, $\boldsymbol{\psi}(\vx, \gZ) = [\va, \vr_1, \vb, \vb, \vr_2, \va, \va, \vr_2, \vc, \vx_1, \ldots, \vx_n]$ and $\boldsymbol{\psi}(\vx, \gZ') =$ $[\vb, \vr_2, \va, \va, \vr_1, \vb, \va, \vr_1, \vc, \vx_1, ..., \vx_n]$. Therefore, it is easy to notice that $\boldsymbol{\psi}(\vx, \gZ) \ne \boldsymbol{\psi}(\vx, \gZ')$, thus the na\"ive encoding is not permutation invariant.

We then show that a na\"ive encoding is not relation inversion invariant. Suppose $\gZ'' = \{(\tta, \ttr_1, \ttb), (\ttb, \ttr_2, \tta), (\ttc, \lnot\ttr_1, \tta)\}$, where $(\tta, \ttr_1, \ttc) \in \gZ$ is changed to its inverse relation $(\ttc, \lnot\ttr_1, \tta)$. Then, $\boldsymbol{\psi}(\vx, \gZ'') = [\va, \vd, \vb, \vb, \ve, \va, \vc, \lnot\vd, \va, \vx_1, \ldots, \vx_n]$ that is different against $\boldsymbol{\psi}(\vx, \gZ)$: $\boldsymbol{\psi}(\vx, \gZ) \ne \boldsymbol{\psi}(\vx, \gZ'')$. Therefore, the na\"ive encoding function is not relation inversion invariant.

In conclusion, from the above two counterexamples, we prove that a na\"ive encoding function $\boldsymbol{\psi}$ is neither permutation invariant nor relation inversion invariant.
\end{proof}

We now provide proof of the permutation invariance and the relation inversion invariance of our invariant and effective graph encoding $\boldsymbol{\psi}^*$, described in Section~3.4.
Before starting the proof, we first revisit the permutation invariant property of graph neural networks that sum, mean and max operators are permutation invariant for the input set of \texttt{AGGR}. Thus, if we use sum, mean, or max for \texttt{AGGR}, then the token embedding perturbation function $\boldsymbol{\beta}$ naturally satisfies the permutation invariance property.
In other words, $\boldsymbol{\beta}(\mX, \gZ) = \boldsymbol{\beta}(\mX, \pi \cdot \gZ)$, where $\mX = \tilde{\boldsymbol{\psi}}(\vx, \texttt{SORT}(\texttt{ENT}(\gZ)))$ for any permutation $\pi$.

\begin{proposition}
\label{prop:ours}
Invariant and efficient encoding $\boldsymbol{\psi}^*$ is both permutation invariant and relation inversion invariant.
\end{proposition}
\vspace{-0.2cm}
\begin{proof}
Suppose $\vx = [x_1, \ldots, x_n]$ and $\gZ = \{(\tta, \ttr_1, \ttb), (\ttb, \ttr_2, \tta), (\tta, \ttr_1, \ttc)\}$. We first consider the permutation invariance for any permuted set $\gZ' = \pi \cdot \gZ$. While $\gZ$ and $\gZ'$ can have different orders of elements thus the outputs of $\texttt{ENT}(\gZ)$ and $\texttt{ENT}(\gZ')$ could be different, we always obtain the same output with the usage of the $\texttt{SORT}$ operator for encoding. In other words, $\texttt{SORT}(\texttt{ENT}(\gZ)) = \texttt{SORT}(\texttt{ENT}(\gZ'))$ holds due to the definition of the $\texttt{SORT}$ operation in Eq.~5 of the main paper.
Therefore, $\tilde{\boldsymbol{\psi}}(\vx, \texttt{SORT}(\texttt{ENT}(\gZ))) = \tilde{\boldsymbol{\psi}}(\vx, \texttt{SORT}(\texttt{ENT}(\gZ')))$ holds.

Further, since the token embedding perturbation function $\boldsymbol{\beta}(\cdot, \gZ)$ along with sum, max, or mean in $\texttt{AGGR}$ is also permutation invariant with regards to any permutation on $\gZ$, we conclude our invariant and efficient encoding $\boldsymbol{\psi}^*$ is permutation invariant.

We finally prove the relation inversion invariance property of $\boldsymbol{\psi}^*$. Suppose $\gZ'' = (\gZ \cup t') \setminus t$ where $t \in \gZ$ is any triplet in a set and $t'$ is inverse of $t$. Then, $\texttt{ENT}(\gZ) = \texttt{ENT}(\gZ'')$ that is trivial as $\texttt{ENT}(\gZ)$ returns the set of only unique nodes in $\gZ$. Therefore, $\tilde{\boldsymbol{\psi}}(\vx, \texttt{SORT}(\texttt{ENT}(\gZ))) = \tilde{\boldsymbol{\psi}}(\vx, \texttt{SORT}(\texttt{ENT}(\gZ'')))$ correspondingly holds. 

The remaining step to conclude the proof is to show the following equality: $\boldsymbol{\beta}(\cdot, \texttt{INV}(\gZ)) = \boldsymbol{\beta}(\cdot, \texttt{INV}(\gZ''))$, to conclude that $\boldsymbol{\psi}^*(\vx, \gZ) = \boldsymbol{\psi}^*(\vx, \gZ'')$ from $\boldsymbol{\beta}( \tilde{\boldsymbol{\psi}}( \vx, \texttt{SORT}(\texttt{ENT}(\mathcal{Z}))), \texttt{INV}(\gZ) ) = \boldsymbol{\beta}( \tilde{\boldsymbol{\psi}}( \vx, \texttt{SORT}(\texttt{ENT}(\mathcal{Z''}))), \texttt{INV}(\gZ'') )$. We note that $\texttt{INV}(\gZ) = \texttt{INV}(\gZ'')$, as $\texttt{INV}$ makes any graph as bidirectional one by the definition in Eq.~6 of the main paper. Therefore, $\boldsymbol{\beta}(\cdot, \texttt{INV}(\gZ)) = \boldsymbol{\beta}(\cdot, \texttt{INV}(\gZ''))$ holds, and the relation inversion invariance property of $\boldsymbol{\psi}^*$ holds.

\end{proof}

\section{Experimental Setup}
\label{appendix:setup}
In this section, we introduce the detailed experimental setups for our models and baselines. Specifically, we describe the details on implementation, dataset, training and model in the following subsections of \ref{appendix:supple:implementation}, \ref{appendix:supple:data}, \ref{appendix:supple:training} and \ref{appendix:supple:model}, one by one.

\subsection{Implementation Details}
\label{appendix:supple:implementation}

We use the T5-small~\citep{T5} as the base Pre-trained Language Model (PLM) for all experiments. For the pre-trained checkpoint, we use the version that the authors released.
For all implementations, we use Pytorch~\citep{Pytorch}. To easily implement the language model, we use the huggingface transformers library~\citep{transformers}.

\paragraph{Retriever Details}
\label{appendix:triplet/encoding}
In this paragraph, we describe the implementation details of our context-relevant subgraph retriever, including the triplet embedding and dialogue context embedding for the retriever.

For the dialogue history embedding function $q$, we use the existing pre-trained language model (PLM). Specifically, we use the encoder part of the T5-small model~\citep{T5} and freeze the parameters of it not to be trained. We then instead add a Multi-Layer Perceptron (MLP) on top of it, to give a point-wise attention~\citep{attention} to each token, whereby all tokens are not equally considered in the sentence encoding. 
Formally,
\begin{gather*}
    q(\vx) = \sum_{i=1}^n \alpha_i * \vz_i, \ 
    \mZ = [\vz_1, \ldots, \vz_n] = \texttt{Enc}(\mX), \\
    \alpha_i = \frac{\exp(\texttt{MLP}(\vz_i))}{\sum_{j=1}^n \exp(\texttt{MLP}(\vz_j))}  \:\forall i
\end{gather*}
where $\alpha_i$ is a scalar, and \texttt{MLP} is a Multi-Layer Perceptron consisting of two linear layers and ReLU nonlinearity.

For obtaining triplet representations, we need to embed the entity (node) and relation (edge) into the latent space.
Similar to the token embedding matrix used in PLMs, we can introduce the entity and relation embedding matrices. However, since the number of entities used in Freebase of OpendialKG~\citep{opendialkg} is too large compared to the number of tokens in T5 (100,814 vs 32,000)~\citep{T5}, it is inefficient to introduce the trainable entity embedding matrix for the retriever. 
Furthermore, the use of standalone entity embedding matrix might be sub-optimal in terms of generalization since there is no evidence that all entities in a large-scale KG emerge in training dataset.

Thus, we instead reuse the contextualized representation from the PLM encoder, to embed each node if the corresponding entity exists in the dialogue context. Formally, suppose that there is a triplet $\{(\tte_h, \ttr, \tte_t)\}$ in the 1-hop subgraph $\gG$, which satisfies the following condition: 
$q(\tte_h) \subseteq \vx$ or $q(\tte_t) \subseteq \vx$.
If so, we can know the position of the mapped entity within the dialogue history: $[x_{start}, ..., x_{end}] = q(\tte_h)$ from $q(\tte_h) \subseteq \vx$.
Therefore, the node embedding for the entity $\tte_h$ is obtained by $\texttt{EntEmb}(\tte_h) = \frac{1}{|q(\tte_h)|} \sum_{i=start}^{end} \texttt{Enc}(\mX)_i \; \text{iff} \;  q(\tte_h) \subseteq \vx$. 
If the entity mention does not exist in the dialogue history, we use the zero vector as the node embedding.
For edge embedding, we use the trainable relation embedding matrix $\mR \in \mathbb{R}^{|\gR| \times 128}$ to represent the edge, since the number of relations is relatively small (1,357).

\begin{figure}
    \centering
    \includegraphics[width=0.5\linewidth]{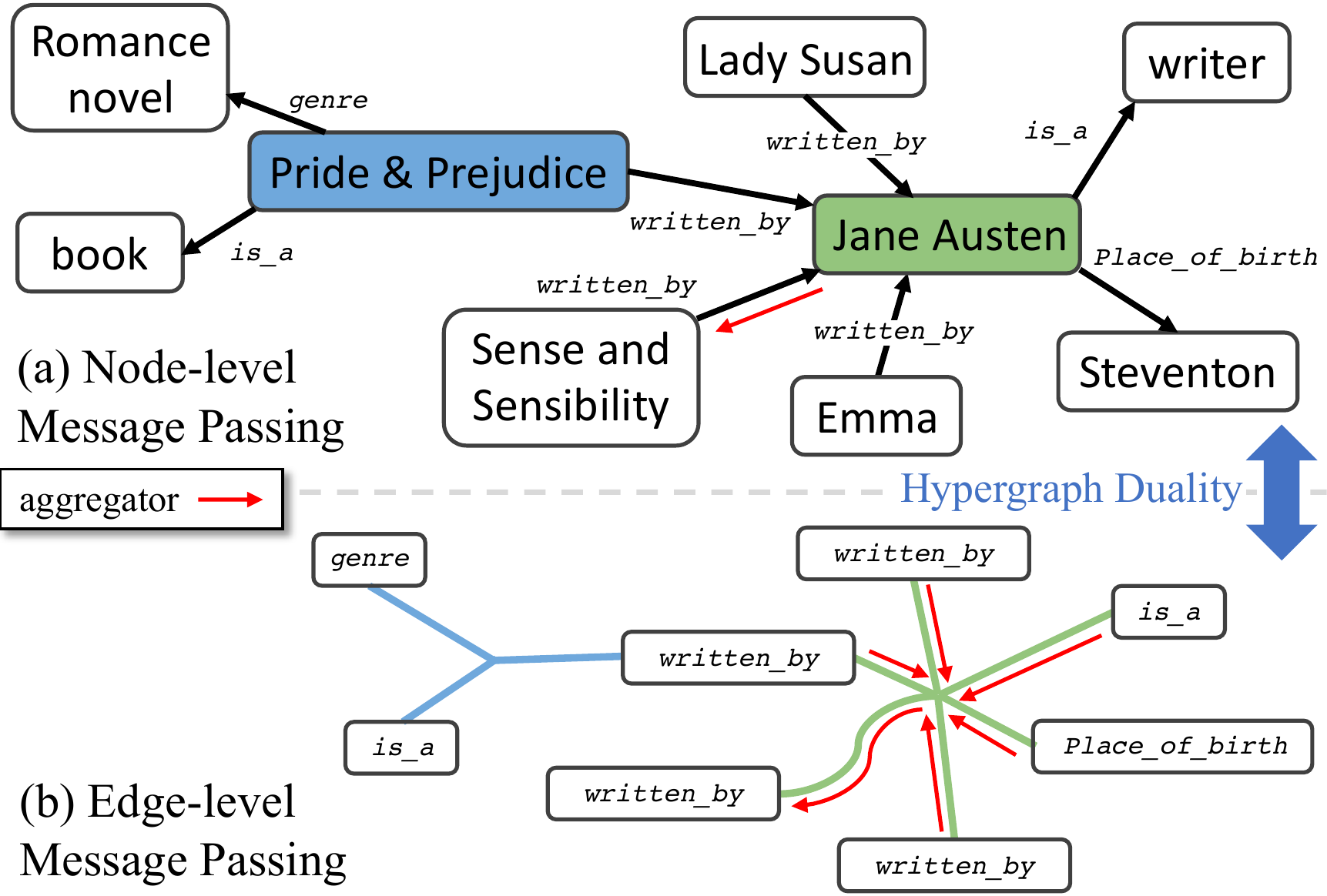}
    \vspace{-0.075in}
    \caption{\small \textbf{GNN-based Triplet Representation for Retrieval.}  To represent each triplet with regards to its graph structure, we use the message passing on both nodes and edges. 
    (a) Node-level Message Passing. To represent the entity \textit{Sense and Sensibility}, the message from its neighbors -- the entity \textit{Jane Austen} -- is aggregated. 
    (b) Edge-level Message Passing. To represent the relation \texttt{written\_by}, the messages from relations associated to a green hyperedge are aggregated. We do not draw self-loops and inverse edges for simplicity.}
    \label{fig:retriever}
    \vspace{-0.15in}
\end{figure}

With our node and edge representations, we now focus on representing the triplet in Eq.~4 of the main paper for its retrieval. In particular, we use the Graph Neural Networks (GNNs) for encoding triplets, where we obtain the node representations from the Graph Convolutional Network (GCN)~\citep{GCN} that is a widely used architecture for representing the nodes with respect to their graph structures. However, for representing the edges, we use the Edge Hypergraph Graph Neural Network (EHGNN) used in~\citet{EHGNN}, due to its simplicity but effectiveness for edge representations. We summarize our triplet representation in~\autoref{fig:retriever}.

\paragraph{Graph Encoding Details}
In this paragraph, we describe the implementation details of the token embedding perturbation function $\boldsymbol{\beta}$ used in our \textit{Invariant and Efficient} graph encoding introduced in Section 3.4. To be aware of the relation of the graph over GNNs, we use the simplified version of CompGCN~\citep{CompGCN}. For architectural details, instead of using the different linear layers to distinguish the inverse relation from its opposite relation, we use the same linear layer. Also, we use subtraction as the specific composition operator for reflecting relations in CompGCN.

\begin{figure*}
    \centering
    \includegraphics[width=0.9\linewidth]{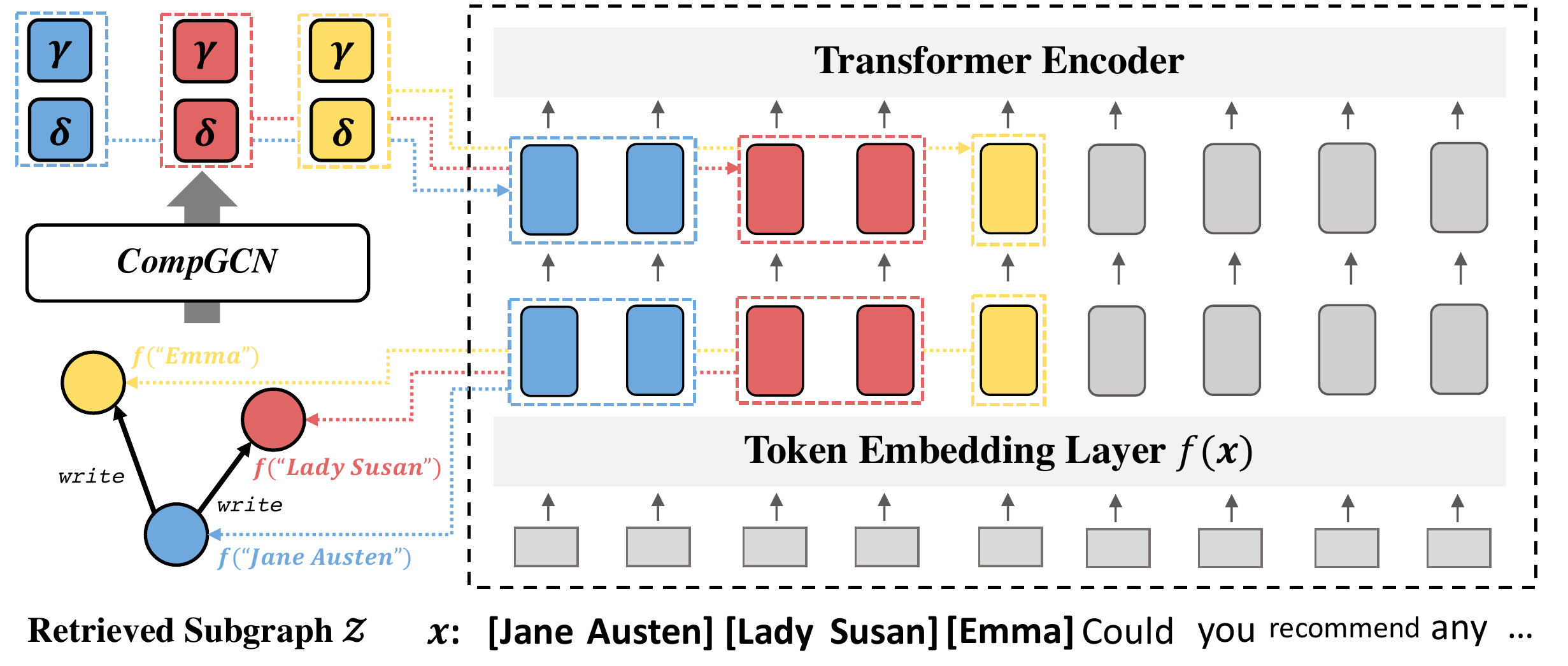}
    \vspace{-0.1in}
    \caption{\small \textbf{Comprehensive diagram for Invariant and Efficient graph encoding.} Our proposed graph encoding first concatenates the sorted list of entities in front of the dialogue history. Then, we form the learnable affine transformation $\gamma, \delta$ for each entity using relation-aware GNN such as CompGCN~\citep{CompGCN}.}
    \label{fig:kfm}
    \vspace{-0.15in}
\end{figure*}

Then, we form the learnable affine transformation based on the aggregated representation from GNN layers, to perturb the token embeddings with respect to their graph information as in Equation 6 of the main paper. In particular,
\begin{equation*}
    \boldsymbol{\eta} = \text{\texttt{R-GNN}}(f(a);\gZ) = \text{\texttt{UPD}}(f(a), \text{\texttt{AGGR}}(\set{f(b), \texttt{r}} {\forall \texttt{b} \in \mathcal{N}(\texttt{a};\gZ)})),
\end{equation*}
\vspace{-0.15in}
\begin{equation*}
    \boldsymbol{\gamma} = \texttt{MLP}_1(\boldsymbol{\eta}), \quad
    \boldsymbol{\delta} = \texttt{MLP}_2(\boldsymbol{\eta}), \quad
    \boldsymbol{\beta}(f(a), \gZ) = (\boldsymbol{1} + \boldsymbol{\gamma}) * f(a) + \boldsymbol{\delta},
\end{equation*}
where $\texttt{MLP}_1$ and $\texttt{MLP}_2$ are learnable MLPs consisting of two linear layers with ReLU nonlinearity.
In~\autoref{fig:kfm}, we illustrate comprehensive diagram of Equation 6, which enables our \textit{Invariant and Efficient} graph encoding to understand the structure of the retrieved subgraph $\gZ$.

\paragraph{Contrastive Learning Details}
For contrastive learning, we initialize $\tau$ in Equation 7 as 0.01.

\paragraph{KQA Details} 
In this paragraph, we describe the implementation details for our Knowledge-verifying Question Answering (KQA) introduced in Section 4.
For building the QA dataset, we first gather the dialogue sessions where the gold response contains the entity from the whole OpendialKG dataset. Then, we extract the triplet from the given whole KG where the head entity is placed within the dialogue history and the tail entity is placed within the gold response. We build a QA training dataset based on the extracted triplets and a corresponding dialogue session. To diversify the training data, we replace the tail entity of each triplet with plausible candidate entities within KG and change the entity in the response following the changed entity on the triplet. As a result, we obtain the QA dataset size of 200k. 
We train the BERT-base~\citep{BERT} with the constructed QA dataset. We hold out 10\% of data for validation and obtain the fine-tuned BERT model with 88.89 F1 score on the hold-out validation set.
When we apply the fine-tuned QA model on the evaluation of the generated responses, we rebuild the QA evaluation set with the generated response instead of a gold response as illustrated in Figure 3 of the main paper.

\subsection{Dataset Details}
\label{appendix:supple:data}
We mainly conduct experiments on \textbf{OpendialKG}~\citep{opendialkg}, which provides the parallel dialogue corpus corresponding to the existing large-scale Knowledge Graph (KG) named Freebase~\citep{Freebase}. The provided large-scale KG consists of total 1,190,658 fact triplets over 100,813 entities and 1,358 relations.
This dataset is collected from 15K human-to-human role-playing dialogues, having multi-turns, from which we pre-process that each assistance response is the label and its corresponding dialogue history is the input.
Although some of the data contain the gold knowledge that is useful for generating the response on the ongoing conversation, we found that $51\%$ of data has no gold knowledge.
To overcome this limitation, we additionally find entities from the dialogue history using the Named Entity Recognition module in spaCy\footnote{https://spacy.io/}, and then include the extracted entities' corresponding triplets in the KG to the dataset.
For entity linking, we use the exact match.
Since the dataset does not provide the pre-defined data split, we randomly split sessions into train (70\%), validation (15\%), and test sets (15\%).
We also conduct experiments on \textbf{KOMOIDS}~\citep{KOMODIS} dataset and follows the same preprocessing as in OpendialKG dataset.

\subsection{Training Details}
\label{appendix:supple:training}
All experiments are constrained to be done with a single 48GB Quadro 8000 GPU. SURGE training needs 12 GPU hours.
For all experiments, we select the best checkpoint on the validation set.
We fine-tune the SURGE for 30 epochs on the training set, where we set the learning rate as 1e-4, weight decay as 0.01, learning rate decay warmup rate as 0.06, maximum sequence length for dialogue history as 256, maximum sequence length for knowledge as 128, and batch size as 24. For retrieval, we use the subgraph size $n$ as 3, and sample size $k$ for marginalization as 4.
We use the AdamW~\citep{AdamW} optimizer for training.
For fair evaluation, we apply the same training setting to all baselines if applicable. All experimental results are reported with a single run.

\subsection{Model \& Baselines Details}
\label{appendix:supple:model}
In this subsection, we describe the details of baselines and our models used in our experiments, as follows:
\vspace{-0.1in}
\begin{enumerate}[itemsep=0.75mm, parsep=0pt, leftmargin=*]
    \item \textbf{No Knowledge}: This model is provided with only the dialog history. No knowledge is used to generate responses.
    
    \item \textbf{Gold Knowledge}: This model is provided with the dialogue history along with its exact gold knowledge for the gold response. Thus, since this model uses such gold knowledge, we expect the results of it as the upper bound of the task.
    
    \item \textbf{Space Efficient (series)}: This model is provided with all the knowledge which are related to the entities that appeared in the dialogue history~\citep{efficientdialogue}, by matching the entities in the dialogue history and the entities in the KG. In particular, this model encodes the entities and their relations explicitly in the words in the encoder part. 
    
    \item \textbf{Space Efficient (parallel)}: This model is mostly the same as the above model -- space Efficient (series) -- except the knowledge encoding part. Specifically, it encodes the entities in the words like the above, whereas, encoding the relation between entities in the segmentation block of the entities~\cite{efficientdialogue}.
    
    \item \textbf{EARL}: This model uses the RNN-based encoder-decoder architecture with the entity-agnostic representation learning~\citep{EARL}, with all the provided knowledge associated with the entities in the dialogue history. Specifically, this model first calculates the probability of words obtained by encoding the entities in the KG, and then uses such probabilities to generate a word in the decoding phase. 
    
    \item \textbf{DiffKG}: This model~\citep{DiffKG} uses a differentiable path reasoning, which is jointly trainable along with the dialogue generation. After the path reasoning, the entities in the reasoning path are naively appended in front of the dialogue history, then concatenated input is forwarded to the pre-trained language model.
    
    \item \textbf{Random Retrieval}: This model is provided with entire facts from k-hop subgraphs of entities that appeared in the dialogue history. However, instead of encoding all the knowledge in one-hop subgraph as in Space Efficient, this model randomly samples them, which are then used for generating responses.
    
    \item \textbf{Sparse Retrieval} (BM25): This model is also provided with entire facts from k-hop subgraphs of entities. To sample relevant facts to the dialogue history among the entire facts, this model uses BM25~\citep{BM25} that is a sparse retrieval model. To be specific, let assume we have a dialogue history and its corresponding facts from k-hop subgraphs of matched entities. Then, to run BM25, we first concatenate components of each fact consisting of two entities and one relation, and tokenize the dialogue history and the facts for obtaining corpus and queries, respectively, for BM25. After that, BM25 calculates the lexical overlapping score between the dialogue context (corpus) and the one-hop fact (query), from which we use the relevant facts having top-$k$ scores by BM25.
    
    \item \textbf{Dense Retrieval} (Bi-encoder, Poly-encoder): This model uses a pre-trained language model for the triplet embedding of the retriever instead of using GNN. Specifically, we consider each triplet as a single sentence (e.g, (Jane Austen, write, Susan) $\rightarrow$ ``Jane Austen write Susan'') and embed them with the pre-trained language model. For scoring, we use both bi-encoder and poly-encoder architectures~\citep{Polyencoder}.
    
   \item \textbf{SURGE (unsupervised)}: Our basic subgraph retrieval-augmented generation framework that is provided with entire facts from k-hop subgraphs of entities. In particular, this model trains the structure-aware subgraph retriever without any guidance of the gold knowledge (i.e., ground truth knowledge for the dialogue history is not given). In other words, for the given dialogue context, this model implicitly learns to retrieve the context-relevant knowledge, and then generates the response with the retrieved knowledge.
   
    \item \textbf{SURGE (semi-supervised)}: Our subgraph retrieval-augmented generation framework with semi-supervised learning of graph retrieval, with provided entire facts from k-hop subgraphs of entities. Unlike the unsupervised version of SURGE, this model trains the retriever to select the gold knowledge if the dialogue context has such knowledge during training.
    
    \item \textbf{SURGE (contrastive)}: Our full subgraph retrieval-augmented generation framework with the contrastive learning of graph-text modalities as well as the semi-supervised learning of graph retrieval, with provided entire facts from k-hop subgraphs of entities. Unlike aforementioned frameworks of ours, this additionally enforces the model to faithfully reflect the retrieved knowledge in the input, to the generated response with contrastive learning. 
\end{enumerate}

\begin{figure*}
    \begin{minipage}{0.38\textwidth}
        \centering
        \includegraphics[width=1.0\linewidth]{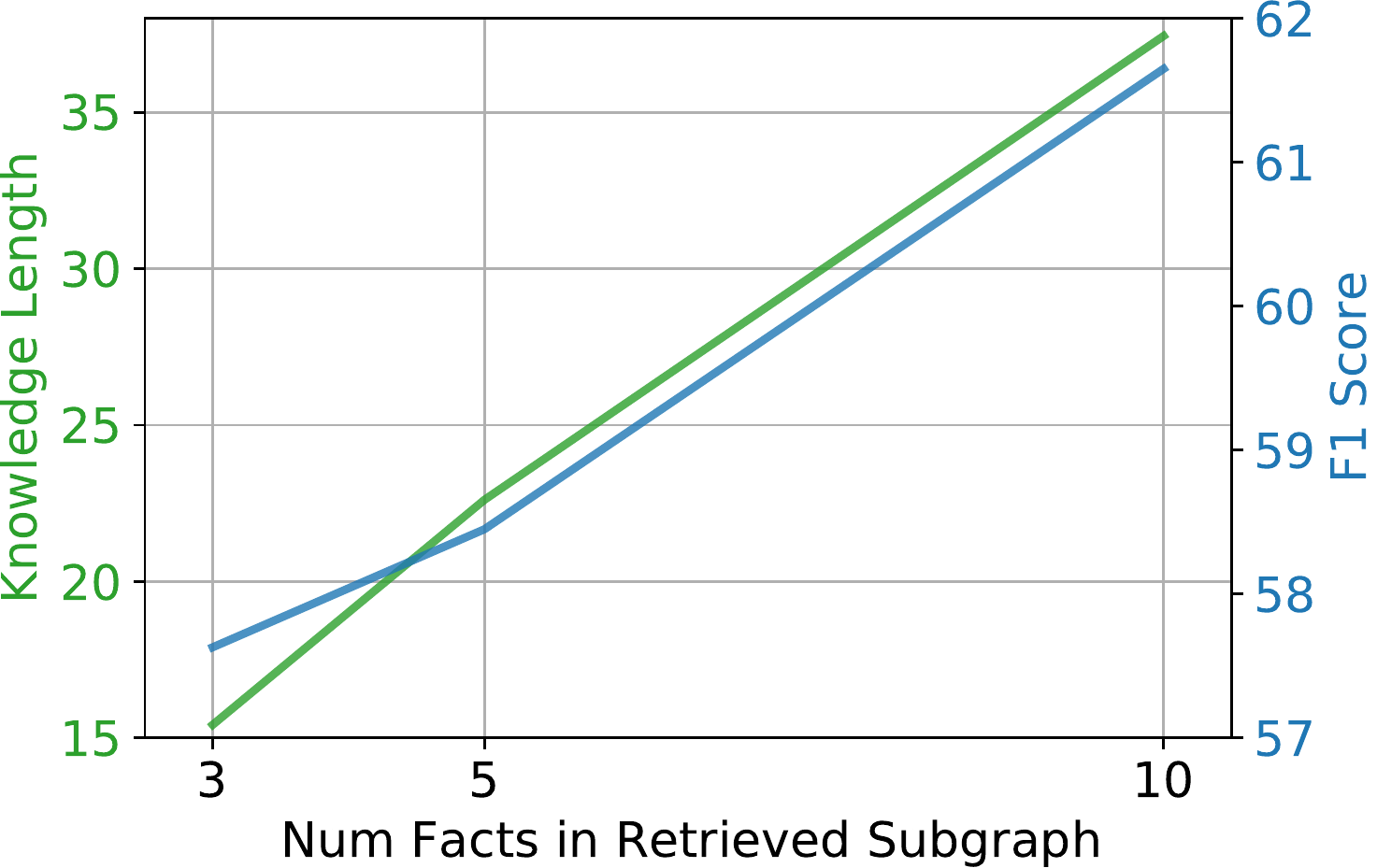}
    \end{minipage}
    \hfill
    \begin{minipage}{0.60\textwidth}
        	\centering
        	\resizebox{1.0\textwidth}{!}{
        	\renewcommand{\arraystretch}{1.5}
        	\renewcommand{\tabcolsep}{1.2mm}
        	\begin{tabular}{lcccccc}
        		\toprule
        		{\textbf{Method}} & MRR & Hits@1 & Hits@3 & Hits@5 & Hits@10 & Hits@100 \\
        		\midrule[0.8pt]
        		\textbf{Random Retrieval} & 8.67 & 3.11 & 7.89 & 10.86 & 17.84 & 66.22 \\
        		\textbf{Sparse Retrieval (BM25)} & 7.76 & 2.61 & 6.72 & 9.89 & 16.67 & 61.21 \\
        		\textbf{Dense Retrieval (Bi-encoder)} & 20.89 & 11.99 & 23.79 & 30.34 & 39.26 & 67.26 \\
        		\textbf{Dense Retrieval (Poly-encoder)} & 21.47 & 12.13 & 24.12 & 31.51 & 41.43 & 68.03 \\
        		\midrule[0.5pt]
        		\textbf{SURGE} (unsupervised) & 21.24 & 10.46 & 24.06 & 31.97 & 44.17 & 74.74 \\
        		\textbf{SURGE} (semi-supervised) & 22.53 & 13.40 & 24.79 & 31.84 & 42.37 & 69.33 \\
         		\textbf{SURGE} (contrastive) & \bf 25.98 & \bf 16.67 & \bf 28.50 & \bf 35.72& \bf 46.11 & \bf 74.31\\
        		\bottomrule
        	\end{tabular}
        	}
    \end{minipage}
    \vspace{-0.05in}
    \caption{\small (Left:) Performances of our SURGE by varying the number of facts for retrieving the subgraph (i.e., varying the number of triplets in the subgraph) from three, to five, to ten, with the length of sequence for knowledge (knowledge length) and F1 scores of KQA as evaluation metrics. (Right:) We additionally report the knowledge retrieval performances, with MRR and Hits@K as evaluation metrics.}
    \label{fig:fact_ablation}
    \vspace{-0.05in}
\end{figure*}
\begin{table*}
	\centering
	\caption{\small Experimental results on OpendialKG dataset with \textbf{BART-base}.}
	\resizebox{0.99\textwidth}{!}{
	\begin{tabular}{lcccccccccc}
		\toprule
    		              & \multicolumn{2}{c|}{\textbf{KQA}} & \multicolumn{4}{c|}{\textbf{BLEU}} & \multicolumn{3}{c|}{\textbf{ROUGE}} & \multicolumn{1}{c}{\textbf{Unigram}}  \\
		{\textbf{Method}} & EM & F1 & B-1 & B-2 & B-3 & B-4 & R-1& R-2& R-L & F1  \\
		\midrule[0.8pt]
		\textbf{No Knowledge} \textit{(BART-base)} & 31.17 & 37.54 & 17.38 & 10.79 & 7.16 & 4.81 & 20.64 & 8.22 & 19.92  & 24.36  \\
		\textbf{Space Efficient} \textit{(BART-base, Series)} & 48.49 & 53.83 & 18.56 & 11.85 & 8.01 & 5.56 & 22.36 & 9.43 & 21.48 & 26.38 \\
		\textbf{Space Efficient} \textit{(BART-base, Parallel)} & 49.80 & 55.06 & \bf18.90 & \bf12.19 & \bf8.35 & \bf5.81 & \bf22.63 & \bf9.79 & \bf21.76 & \bf26.79 \\
		\midrule
		\textbf{SURGE} \textit{(BART-base, semi-supervised, $n=10$)}  & 50.84 & 57.35 & 17.80 & 11.12 & 7.48 & 5.18 & 18.64 & 7.27 & 17.77 & 22.07 \\
		\textbf{SURGE} \textit{(T5-small, semi-supervised, $n=3$)}  & 51.32 & 58.45 & 17.63 & 11.28 & 7.41 & 5.39 & 21.74 & 9.18 & 20.85 & 25.57 \\
		\textbf{SURGE} \textit{(T5-small, semi-supervised, $n=10$)}  & 
		\textbf{54.50} & \textbf{61.65} & 17.70 & 11.37 & 7.81 & 5.50 & 21.55 & 9.09 & 20.65 & 25.44 \\
		\bottomrule
	\end{tabular}
	}
	\label{appendix:bart_exp}
\end{table*}

\section{Additional Experiments}
\label{appendix:additional_exps}

\subsection{Varying the Number of Facts in Subgraphs}
We experiment our SURGE framework with varying the number of facts in retrieval, which are then used in our graph encoding function to condition the encoded graph information for response generation. Specifically, in~\autoref{fig:fact_ablation}, we report the length of sequence for knowledge (knowledge length) and F1 scores measured by our KQA for our SURGE framework, with different numbers of facts within a retrieved subgraph: $n = [3, 5, 10]$. Note that, in this experiment, we only use the semi-supervised model without the contrastive loss.
We expect that the performance of our SURGE will increase as we increase the number of facts within the retrieved subgraph, since the model can leverage more numbers of knowledge for response generation. As shown in~\autoref{fig:fact_ablation}, we observe the significant performance improvements on using ten facts against using three and five facts, while the performance difference between the three and five is marginal. We suggest that this result should be interpreted with the retrieval results on the right side of~\autoref{fig:fact_ablation}, where about 40\% of retrieved subgraphs including the ten different facts contain at least one necessary knowledge, thus the generation performance is boosted according to the improvement in retrieval.

\subsection{Discussions on Using Larger PLMs}
Notably, we observe that the use of larger Pre-trained Language Models (PLMs) -- three times more number of parameters compared to T5-small that we use -- does not result in better performance for the knowledge-grounded dialogue task.
Specifically, in~\autoref{appendix:bart_exp}, we report the experimental results of selected baselines and our SURGE semi-supervised model with BART-base~\citep{BART} as the base PLM.
We want to clarify that the BART-base model has 220M parameters, which is about \textbf{three times larger} than the number of parameters of the T5-small model (60M).

We first observe that BART-base shows decent performance without any knowledge (No Knowledge) compared to the no-knowledge case of T5-small, verifying that the larger PLM generally contains more factual knowledge within its pre-trained parameters. Moreover, BART-base obtains higher scores in the simple word overlap metrics such as BLEU ~\citep{BLEU} and ROUGE~\citep{ROUGE}, whose results further confirm that a larger PLM can generate more natural or syntactically better sentences than the smaller one, thanks to its parameter size. 

On the other hand, we find that BART-base is less suffered from the irrelevant knowledge issue (i.e., conditioning irrelevant knowledge for the given context when generating responses) than T5-small, therefore, the performance of \textit{Space Efficient Encoding} on KQA is quite high. However, the use of BART-base does not result in significant improvement on the KQA metric for our SURGE framework. Moreover, ours with T5-small shows better performance than ours with BART-base in terms of KQA scores, when the number of facts within the retrieved subgraph is 10: $n=10$. This result suggests that the quality of the generated response -- having relevant knowledge to the given context -- might depend on the performance of the subgraph retriever whose goal is to retrieve the context-relevant knowledge, rather than the inherent performance of PLMs.

\begin{table*}
	\centering
	\caption{\small Experimental results on KOMODIS dataset with T5-small.}
	\resizebox{0.99\textwidth}{!}{
	\begin{tabular}{lcccccccccc}
		\toprule
    		              & \multicolumn{2}{c|}{\textbf{KQA}} & \multicolumn{4}{c|}{\textbf{BLEU}} & \multicolumn{3}{c|}{\textbf{ROUGE}} & \multicolumn{1}{c}{\textbf{Unigram}}  \\
		{\textbf{Method}} & EM & F1 & B-1 & B-2 & B-3 & B-4 & R-1& R-2& R-L & F1  \\
		\midrule[0.8pt]
		\textbf{Random} & 12.41 & 14.17 & 7.74 & 4.02 & 2.46 & 1.68 & 21.79 & 4.00 & 21.44  & 16.29  \\
		\textbf{Space Efficient} \textit{(Series)} & 12.41 & 14.70 & 8.34 & 5.13 & 3.77 & 3.05 & 22.36 & 4.85 & 22.06 & 17.37 \\
		\textbf{Space Efficient} \textit{(Parallel)} & 16.46 & 18.70 & 9.33 & 5.66 & 4.06 & 3.20 & 22.80 & 4.12 & 22.47 & 17.72 \\
		\midrule
		\textbf{SURGE} \textit{(unsupervised)}  & 
		16.18 & 18.51 & 11.46 & 7.10 & 5.15 & 4.07 & 23.49 & 5.77 & 23.09 & 18.70 \\
		\textbf{SURGE} \textit{(semi-supervised)}  & 
		16.62 & 19.48 & 11.28 &  6.98 & 5.05 & 3.98 & 23.58 & 5.79 & 23.21 & 18.68 \\
		\textbf{SURGE} \textit{(contrastive)}  & 
		\bf 17.30 & \bf 19.50 & \bf 11.51 & \bf 7.18 & \bf 5.20 & \bf 4.10 & \bf 24.13 & \bf 6.17 & \bf 23.74 & \bf 19.51 \\
		\bottomrule
	\end{tabular}
	}
	\label{tab:appendix:komodis}
\end{table*}

\subsection{Full Experimental Results on KOMODIS}
\label{appendix:exp:komodis}
In the main paper, we mostly focus on OpendialKG dataset~\citep{opendialkg}, since it is the largest and most realistic public datasets that provides both dialogues across diverse domains and corresponding large-scale Knowledge Graph (KG)~\citep{Freebase}.
To verify the effectiveness of our SURGE framework, the existence of the large-scale KG and the importance of relevant fact searching is important since we focus on the real-world scenario where the response generation requires the relevant fact acquirement from the large-scale KG.

However, one can raise the question regarding the  versatility of our method on other datasets. To alleviate the issue, we conduct additional experiments on another dataset named KOMODIS~\citep{KOMODIS}, which is also KG-based dialogue dataset. Compared to OpendialKG, KOMODIS does not provide the corresponding large-scale KG and most of responses do not require the knowledge. Therefore, we only measure the automatic evaluation to evaluate the performance of each method on KOMODIS dataset. In~\autoref{tab:appendix:komodis}, we present the experimental results on the KOMODIS dataset. Results obviously show that our SURGE framework shows superior performance against baselines on the additional dataset. Therefore, we can conclude that our method can generalize to other datasets beyond the opendialKG dataset.

\subsection{Diversity Evaluation}
In the main paper, we evaluate model generation performance primarily on its quality. We measure the distinct metric~\citep{distinct}, which is one of the most popular metrics for evaluating the diversity of the generative model, to evaluate the performance of each model in more diverse aspects.
In~\autoref{tab:rebuttal} left, we report the performance of baselines and our models in distinct metric. Our SURGE framework generates more diverse responses than all other baselines, according to the results.

\subsection{Ablations Studies on GNN Design Choices}
We use two different types of Graph Neural Networks (GNN) in our SURGE framework. One is the Graph Convolutional Network (GCN)~\citep{GCN}, which is used to embed each node entity on the entire 1-hop subgraph in the triplet embedding function $d$ of the main paper Equation 4.
Another is Composition-Based Multi-Relational Graph Convolutional Networks (CompGCN)~\citep{CompGCN}, which is used to embed each entity by considering the relations between entities in the token embedding perturbation function $\boldsymbol{\beta}$ of the main paper Equation 6.
In this subsection, we conduct ablation studies on both GNN design choices.
First of all, we replace the GCN in Equation 4 with Graph Attention Network (GAT)~\citep{GAT} to validate the effect of the GNN design choices on the node embedding in the triplet embedding function.
Then, we run experiments by changing CompGCN in Equation 6 to GCN to see how important the relationships are in the graph encoding.
We present the results on~\autoref{tab:rebuttal} right. Results indicate that the use of GAT in Equation 3 does not have any impact on the performance a lot.
However, the use of relation-aware GNN is highly important in effective and efficient graph encoding, since removing the relation awareness of GNN reduces the performance of our model a lot.

\subsection{Automatic Evaluations on Knowledge Groundedness}
\label{appendix:exp:groundedness}
Our KQA metric introduced in Section 4 is useful to evaluate the knowledge groundedness of the generated response. 
Since the novelty of the KQA metric stems from the use of the KG to resolve the issue from the missing knowledge by only considering the gold response for evaluation, we can also utilize other rule-based metrics like \textbf{string matching} (check whether at least one of the answer candidates of KQA presents in the generated response by string matching) or \textbf{Entity F1} score (measuring the F1 score against each entity in answer candidates instead of the gold response).
As we all know, automatic evaluations can be imperfect when compared to human evaluations. However, we also believe that using a variety of credible automatic evaluation metrics will strengthen the validity of the experimental results. Therefore, we supplement the experimental results with three more evaluation metrics for measuring whether the generated responses contain appropriate knowledge.

In~\autoref{tab:appendix:knowledge}, we measure Knowledge F1 (KF1 in Table 3 in the main paper), string matching, and entity F1 for representative baselines and our SURGE (semi-supervised) in OpendialKG, as an extension of Table 1 in the main paper. For KF1, we measure the F1 score regarding the concatenation of the question (head entity and relation) and all answer candidates (available tail entities) in KQA as the gold response. The results show that all metrics show the same tendency with KQA and our proposed method still outperforms other baselines by generating responses with more proper knowledge.
Although three rule-based metrics are useful for assessing the knowledge groundedness of generated responses, they do have some drawbacks. KF1 and Entity F1 are affected by the length of the generated response and answer candidates. String matching is too strict since it may miss some responses that only contain partial words of knowledge (e.g., the response only contains the first name of the author whereas the candidate answers contain the full name of the author). As a result, the use of KQA is also beneficial since the trained QA model can compensate for the shortcomings of rule-based metrics.

\begin{figure*}
    \captionsetup{type=table}
    \captionof{table}{\small (Left:) Performance evaluation with the diversity metric named Distinct. (Right:) Ablation study results on GNN variants in our modules.}
    \begin{minipage}{0.23\textwidth}
        \centering
        	\resizebox{1.0\textwidth}{!}{
        	\renewcommand{\arraystretch}{1.5}
        	\renewcommand{\tabcolsep}{1.2mm}
        	\begin{tabular}{lcc}
        		\toprule
        		{\textbf{Method}} & Dist-1 & Dist-2 \\
        		\midrule[0.8pt]
        		\textbf{No Knowledge} & 6.06 & 15.73 \\
        		\textbf{All Knowledge} & 9.67 & 24.45 \\
        		\textbf{SEE (Series)} & 8.49 & 21.77 \\
        		\textbf{SEE (Parallel)} & 8.78 & 22.70 \\
        		\textbf{EARL} & 5.15 & 16.46 \\
        		\textbf{Sparse Retrieval (BM25)} & 7.65 & 19.63 \\
        		\midrule[0.5pt]
        		\textbf{SURGE} (semi-supervised) & \textbf{10.18} & \textbf{27.85}  \\
        		\bottomrule
        	\end{tabular}
        	}
    \end{minipage}
    \hfill
    \begin{minipage}{0.75\textwidth}
    	\centering
    	\resizebox{1.0\textwidth}{!}{
    	\renewcommand{\arraystretch}{1.5}
    	\renewcommand{\tabcolsep}{1.2mm}
    	\begin{tabular}{clcccccccccc}
    		\toprule
    		& & \multicolumn{2}{c|}{\textbf{KQA}} & \multicolumn{4}{c|}{\textbf{BLEU}} & \multicolumn{3}{c|}{\textbf{ROUGE}} & \multicolumn{1}{c}{\textbf{Unigram}}  \\
		    & {\textbf{Method}} & EM & F1 & B-1 & B-2 & B-3 & B-4 & R-1& R-2& R-L & F1  \\
    		\midrule[0.8pt]
    		& Eq 4. GCN $\rightarrow$ GAT  & 49.16 & 56.10 & 17.42 & 10.96 & 7.39 & 5.17 & 21.10 & 8.65 & 20.25 & 24.79 \\
    		& Eq 7. CompGCN $\rightarrow$ GCN  & 48.61 & 55.53 & 17.48 & 10.97 & 7.34 & 5.05 & 21.23 & 8.73 & 20.37 & 24.77 \\
    		\midrule[0.5pt]
    		& \textbf{SURGE} \textit{(semi-supervised)}  & \textbf{51.00} & \textbf{57.63} & \textbf{17.70} & \textbf{11.21} & \textbf{7.61} & \textbf{5.28} & \bf21.43 & \bf8.85 & \bf20.57 & \bf25.07 \\
    		\bottomrule
    	\end{tabular}
    	}
    \end{minipage}
    \vspace{-0.05in}
    \label{tab:rebuttal}
    \vspace{-0.05in}
\end{figure*}

\begin{table*}
	\centering
	\caption{\small Experimental results on OpendialKG with additional three metrics other than KQA for measuring whether the generated responses contain appropriate knowledge.}
	\resizebox{0.85\textwidth}{!}{
	\begin{tabular}{lcccccccccc}
		\toprule
    		              & \multicolumn{2}{c|}{\textbf{KQA}} &  &  &   \\
		{\textbf{Method}} & EM & F1 & \textbf{Knowledge F1} & \textbf{Entity F1} & \textbf{String Matching} \\
		\midrule[0.8pt]
            \textbf{No Knowledge} & 12.25 & 20.69 & 13.80 & 9.33 & 13.03 \\
		\textbf{Random Knowledge} & 31.72 & 38.95 & 16.29 & 16.49 & 32.71 \\
            \textbf{All Knowledge} & 43.58 & 50.60 & 18.91 & 21.10 & 44.25 \\
		\textbf{Space Efficient} \textit{(Parallel)} & 38.54 & 44.34 & 17.43 & 18.93 & 40.56 \\
            \textbf{Dense Retrieval} \textit{(Poly-encoder)} & 46.05 & 52.57 & 19.72 & 21.46 & 48.41 \\
            \textbf{DiffKG} & 12.25 & 20.99 & 14.44 & 9.37 & 13.23 \\
		\midrule
		\textbf{SURGE} \textit{(ours, semi-supervised)}  & \bf 51.00 & \bf 57.63 & \bf 21.87 & \bf 23.03 & \bf 55.79 \\
            \midrule
		\textbf{Gold Response} \textit{(oracle)}  & 93.30 & 95.21 & 28.62 & 29.06 & 85.75 \\
		\bottomrule
	\end{tabular}
	}
	\label{tab:appendix:knowledge}
\end{table*}

\begin{figure*}
    \centering
    \includegraphics[width=0.6\linewidth]{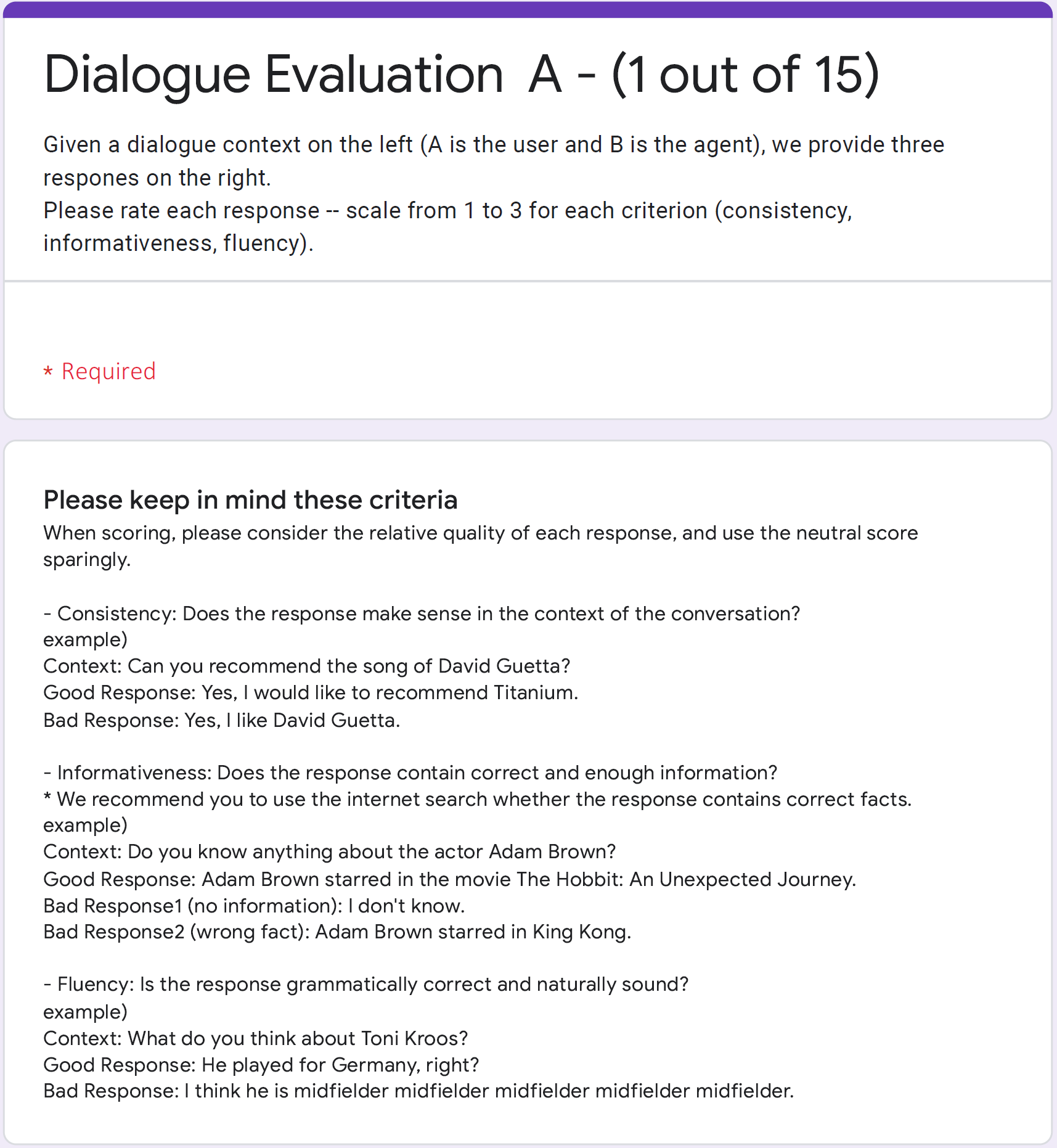}
    \vspace{-0.075in}
    \caption{\small \textbf{Human Evaluation Instructions.} To measure the qualitative performances of the generated responses, annotators are provided with the following instruction on three criteria -- consistency, informativeness, and fluency.
    }
    \label{fig:instruction}
    \vspace{-0.15in}
\end{figure*}

\section{Human Evaluation}
In this section, we describe the details of human evaluation used in Table 7 of the main paper. 
We request the annotators to evaluate the responses generated from two baselines (i.e., ALL Knowledge and Space Efficient) and our SURGE framework in response to the given dialogue context, according to three criteria -- consistency, informativeness, and fluency. 
\autoref{fig:instruction} is the instructions provided to each annotator.
Specifically, regarding the consistency metric, we ask annotators to check whether the generated response makes sense in the context of the conversation. 
For informativeness, we ask annotators to check whether the response contains correct and enough information, whereby experiment participants are recommended to use the internet search, to check whether the response contains correct facts. In addition to this, we also provide the dialogue-related facts from Freebase as a reference for fact checking for annotators.
For fluency, we ask annotators to check whether the response is grammatically correct and naturally sound.

\section{Retrieval and Generation Examples}
\label{appendix:examples}
In this section, we provide examples for knowledge retrieval and response generation, for the given dialogue history.

\begin{figure*}
    \centering
    \includegraphics[width=0.8\linewidth]{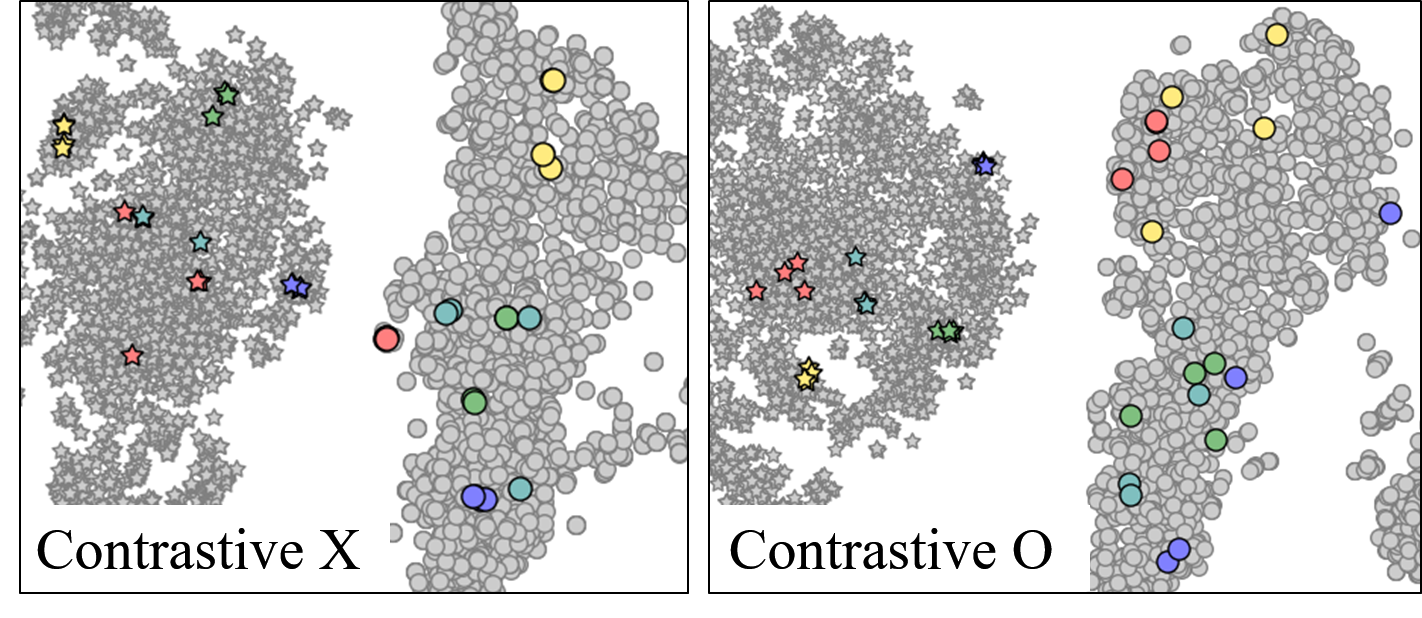}
    \vspace{-0.075in}
    \caption{\small Large version of Figure 5 in the main paper. Stars indicate the embedding of graph and circles indicate the embedding of decoder hidden states (text), respectively.}
    \label{fig:contrastive_big}
    \vspace{-0.15in}
\end{figure*}

\paragraph{Embedding Space Visualization}
In~\autoref{fig:contrastive_big}, we present a larger version of Figure 5 in the main paper.
Specifically, we embed the hidden representations before the projection layer for each graph (star) and the embedding of the generated text (circle) through the dimensionality reduction using t-SNE~\citep{tSNE}. As mentioned in the main paper, the visualization highlights that our SURGE framework with graph-text contrastive learning generates more distinct responses to different subgraphs, unlike the one without graph-text contrastive learning which shows less variety over responses even with different graphs.

\begin{figure*}[!b]
    \centering
    \includegraphics[width=0.9\linewidth]{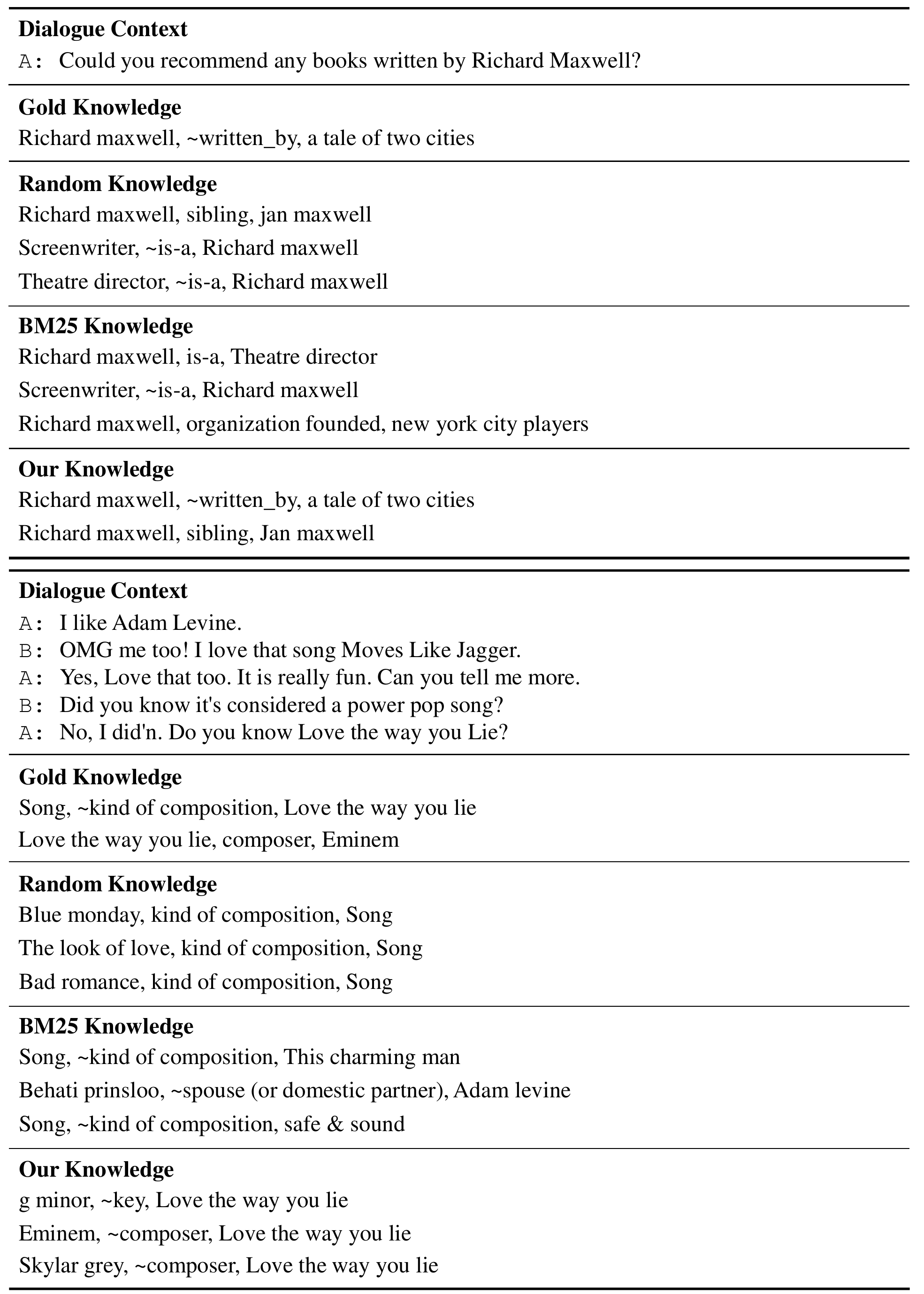}
    \vskip -0.05in
    \caption{\small Examples of the dialogue history with its corresponding gold knowledge as well as the retrieved knowledge from random retrieval and sparse retrieval baselines and from our SURGE framework. The retrieved fact is represented  as the format of (head, relation, tail), where $\sim$symbol in the front of relation (i.e., $\sim$relation) in the retrieved knowledge denotes the inverse relation.}
    \vskip -0.2in
    \label{fig:suppl/retrieval/example}
\end{figure*}

\paragraph{Retrieval Examples} We provide the retrieval examples of various models, such as random retrieval, sparse retrieval and our SURGE models. In particular, in the first (top) example of~\autoref{fig:suppl/retrieval/example}, we are given a dialogue context in regard to books for Richard Maxwell, and baselines including random and BM25 retrievers select the facts associated to the entity Richard Maxwell, which are but irrelevant to the ongoing conversion, for example, (Richard maxwell, is-a Theatre director). Also, as shown in the second (bottom) example of~\autoref{fig:suppl/retrieval/example}, we observe that the simple term-based matching model (i.e., BM25) cannot contextualize the current and previous dialogues, but retrieves the facts associated to frequent words, for example, song, which are less meaningful for the user's question. In contrast to baselines, as our SURGE framework trains a retriever in an end-to-end fashion, it first contextualizes the given dialogue context, and then accurately retrieves relevant knowledge.

\begin{figure*}[!b]
    \centering
    \includegraphics[width=0.9\linewidth]{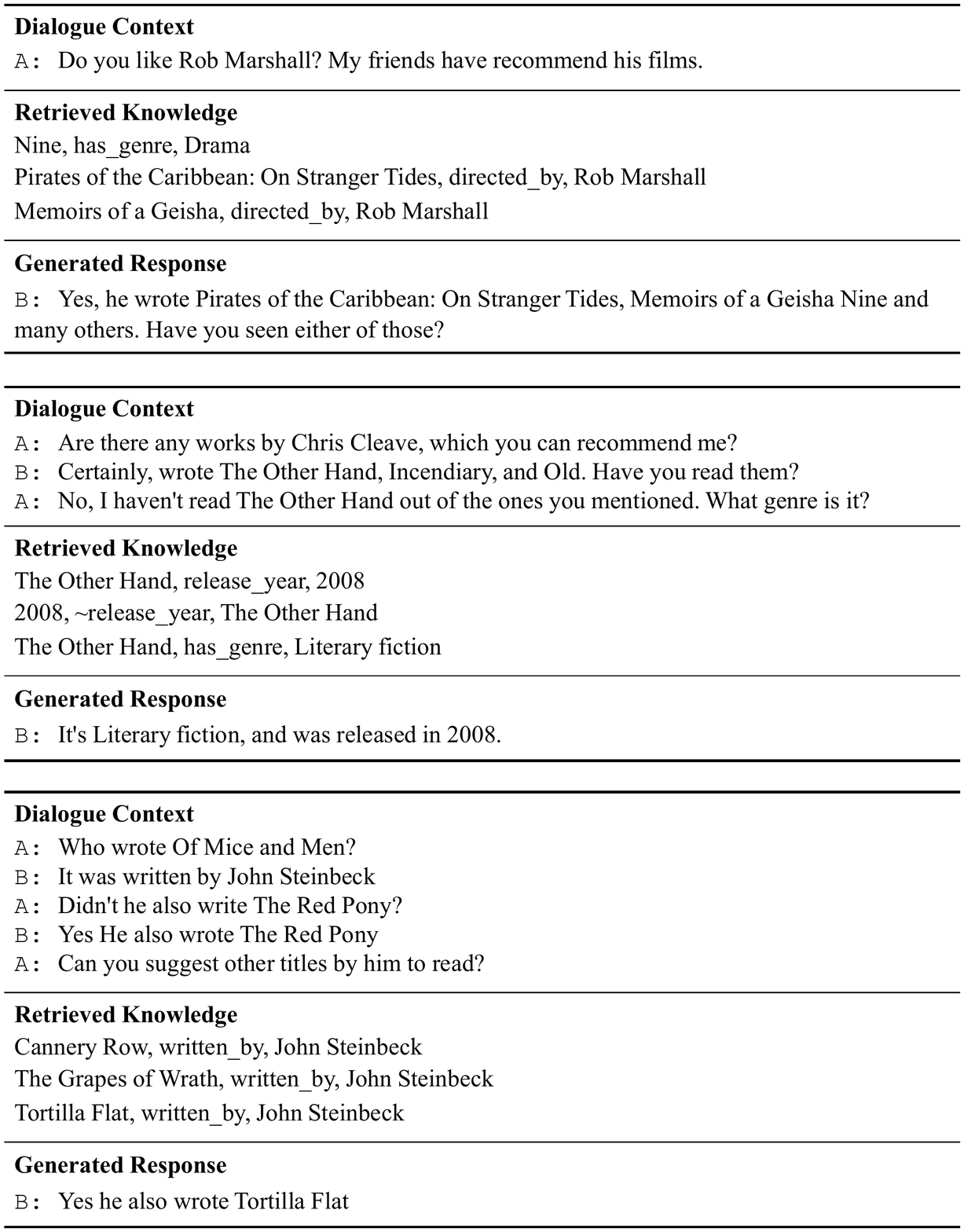}
    \vskip -0.2in
    \caption{\small Examples of the dialogue history with its corresponding retrieved knowledge and generated response from our SURGE framework. The fact is represented  as the format of (head, relation, tail), where $\sim$symbol in the front of relation (i.e., $\sim$relation) in the retrieved knowledge denotes the inverse relation. In this example, we only provide the correct cases of both retrieval and generation.}
    \vskip -0.2in
    \label{fig:suppl/example1}
\end{figure*}

\begin{figure*}[!b]
    \centering
    \includegraphics[width=0.9\linewidth]{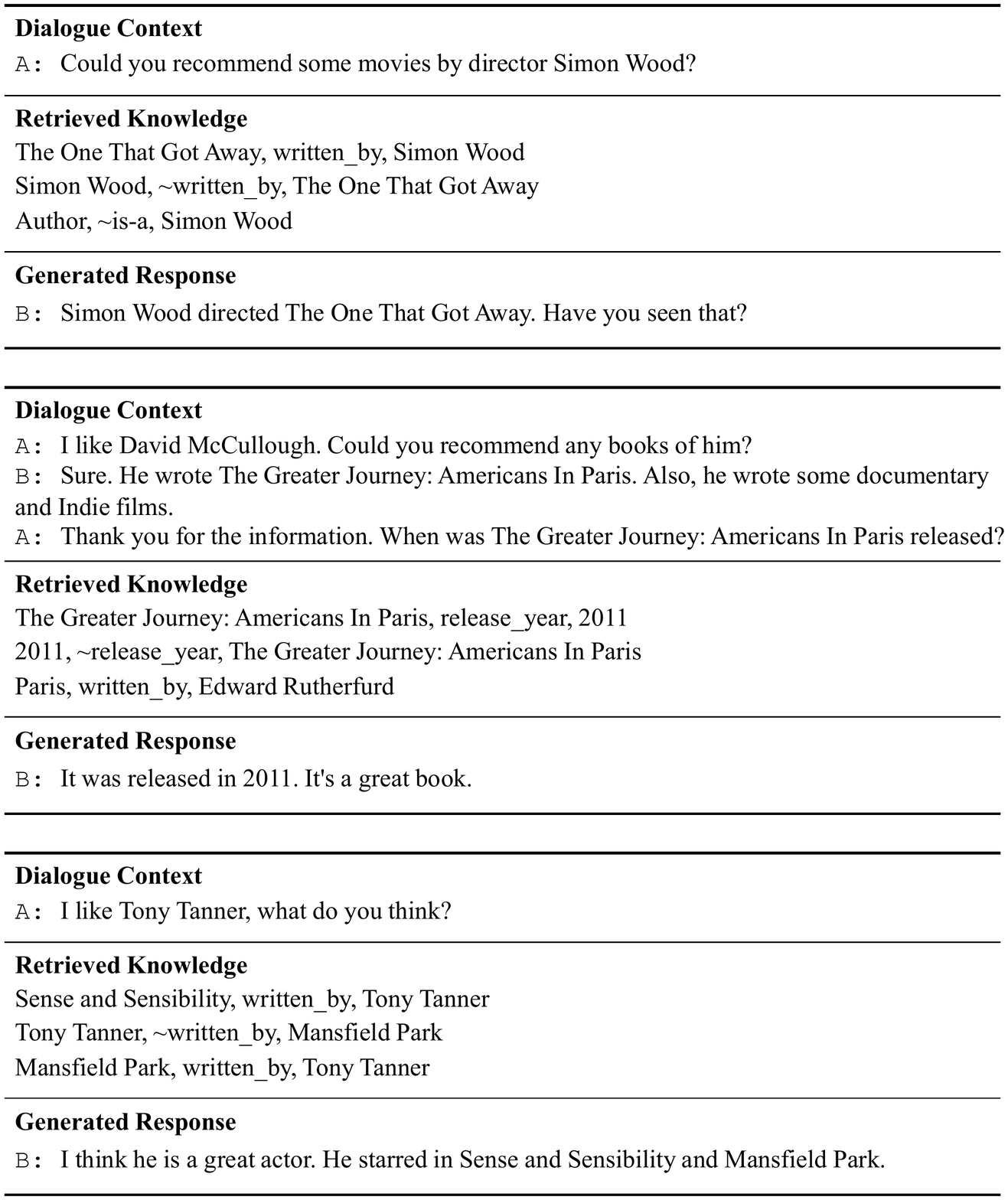}
    \vskip -0.2in
    \caption{\small Examples of the dialogue history with its corresponding retrieved knowledge and generated response from our SURGE framework. The fact is represented  as the format of (head, relation, tail), where $\sim$symbol in the front of relation (i.e., $\sim$relation) in the retrieved knowledge denotes the inverse relation. In this example, we only provide the correct cases of both retrieval and generation.}
    \vskip -0.2in
    \label{fig:suppl/example2}
\end{figure*}

\begin{figure*}[!b]
    \centering
    \includegraphics[width=0.9\linewidth]{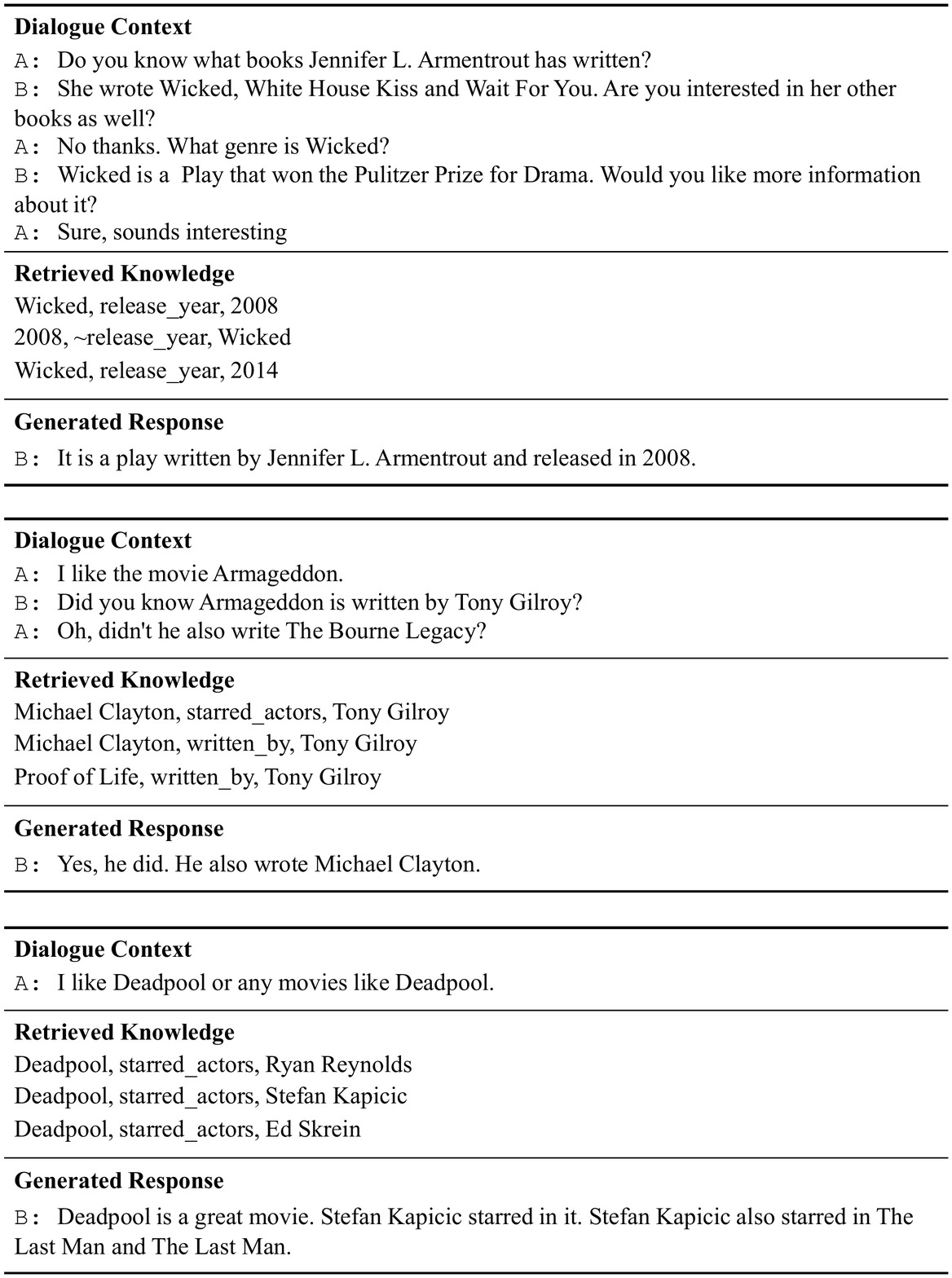}
    \vskip -0.2in
    \caption{\small Examples of the dialogue history with its corresponding retrieved knowledge and generated response from our SURGE framework. The fact is represented  as the format of (head, relation, tail), where $\sim$symbol in the front of relation (i.e., $\sim$relation) in the retrieved knowledge denotes the inverse relation. In this example, we only provide the failure cases due to the problem on data (first row), retrieval (second row), and generation (third row).}
    \vskip -0.2in
    \label{fig:suppl/example3}
\end{figure*}

\paragraph{Generation Examples} We provide the generation examples from our model. To be specific, we provide the dialogue context along with its corresponding retrieved subgraph and generated response obtained from our SURGE framework. In~\autoref{fig:suppl/example1} and~\autoref{fig:suppl/example2}, we provide the correct examples: our model retrieves a context-relevant subgraph, but also generates a factual response from retrieved knowledge. On the other hand, in~\autoref{fig:suppl/example3}, we provide the failure cases. In particular, as shown in the first row of~\autoref{fig:suppl/example3}, the fact in the knowledge graph could be ambiguous or inaccurate, as it defines the release year of the book -- Wicked -- as both 2008 and 2014. Moreover, we further provide the failure example on retrieval in the second row of~\autoref{fig:suppl/example3}, where the user asks about the Bourne Legacy, while the dialogue agents retrieve the irrelevant knowledge to the question. Finally, we show the common problem in PLMs in the last row of~\autoref{fig:suppl/example3}, where the generative model repeats the meaningless words at the end, while the retriever correctly selects the relevant knowledge.

\end{document}